\documentclass[sigconf]{acmart}

\AtBeginDocument{%
  \providecommand\BibTeX{{%
    \normalfont B\kern-0.5em{\scshape i\kern-0.25em b}\kern-0.8em\TeX}}}

\setcopyright{acmcopyright}
\copyrightyear{2021}
\acmYear{2021}
\acmDOI{10.1145/1122445.1122456}

\acmConference[Woodstock '18]{Woodstock '18: ACM Symposium on Neural
  Gaze Detection}{June 03--05, 2018}{Woodstock, NY}
\acmBooktitle{Woodstock '18: ACM Symposium on Neural Gaze Detection,
  June 03--05, 2018, Woodstock, NY}
\acmPrice{15.00}
\acmISBN{978-1-4503-XXXX-X/18/06}

\usepackage{multirow}
\usepackage{footmisc}
\usepackage{subcaption,siunitx,booktabs}
\usepackage{caption}
\usepackage{bbm}
\usepackage{xcolor}
\usepackage[ruled]{algorithm2e}
\usepackage{balance}
\usepackage{enumitem}

\SetKwInput{KwInput}{Input}                
\SetKwInput{KwOutput}{Output}              

\newtheorem{problem}{Problem}
\newtheorem{hypothesis}{Hypothesis}

\newcommand\subfigwidth{0.33}

\begin{document}

\title{Stable Prediction on Graphs with Agnostic Distribution Shift}

\renewcommand{\shortauthors}{Trovato and Tobin, et al.}
\newcommand{\etal}{\textit{et al}.}
\newcommand{\ie}{\textit{i}.\textit{e}.}
\newcommand{\eg}{\textit{e}.\textit{g}.}
\newcommand{\vpara}[1]{\vspace{0.05in}\noindent\textbf{#1 }}

\author[S. Zhang, K. Kuang, J. Qiu, J. Yu, Z. Zhao, H. Yang, Z. Zhang, F. Wu]{
	Shengyu Zhang$^{1}$, Kun Kuang$^{1}$, Jiezhong Qiu$^{2}$, Jin Yu$^{3}$, Zhou Zhao$^{1}$, Hongxia Yang$^{3}$, Zhongfei Zhang$^{1}$ and Fei Wu$^{1}$
}
\affiliation{
    $^1$ College of Computer Science and Technology, Zhejiang University \country{China} 
}
\affiliation{
	$^2$ Department of Computer Science and Technology, Tsinghua University \country{China} 
}
\affiliation{
	$^3$ Alibaba Group \country{China} 
}

\email{
  {sy_zhang, kunkuang, zhaozhou, wufei}@zju.edu.cn
}
\email{
  {kola.yu, yang.yhx}@alibaba-inc.com
}
\email{
	qiujz16@mails.tsinghua.edu.cn
}
\renewcommand{\authors}{Shengyu Zhang, Tan Jiang, Tan Wang, Kun Kuang, Zhou Zhao, Jianke Zhu, Jin Yu, Hongxia Yang, Fei Wu}



\begin{abstract}
Graph is a flexible and effective tool to represent complex structures in practice and graph neural networks (GNNs) have been shown to be effective on various graph tasks with randomly separated training and testing data. In real applications, however, the distribution of training graph might be different from that of the test one (\eg, users' interactions on the user-item training graph and their actual preference on items, \ie, testing environment, are known to have inconsistencies in recommender systems). Moreover, the distribution of test data is always agnostic when GNNs are trained. Hence, we are facing the agnostic distribution shift between training and testing on graph learning, which would lead to unstable inference of traditional GNNs across different test environments. To address this problem, we propose a novel stable prediction framework for GNNs, which permits both locally and globally stable learning and prediction on graphs. 
In particular, since each node is partially represented by its neighbors in GNNs, we propose to capture the stable properties for each node (locally stable) by re-weighting the information propagation/aggregation processes. For global stability, we propose a stable regularizer that reduces the training losses on heterogeneous environments and thus warping the GNNs to generalize well. We conduct extensive experiments on several graph benchmarks and a noisy industrial recommendation dataset that is collected from 5 consecutive days during a product promotion festival. The results demonstrate that our method outperforms various SOTA GNNs for stable prediction on graphs with agnostic distribution shift, including shift caused by node labels and attributes.
\end{abstract}



\maketitle

\section{Introduction}

Graphs are commonly used to represent the complex structures of interacting entities in a flexible and effective manner, with applications and domains varying from social networks, knowledge graphs, and recommender systems. The challenging and open-ended nature of graph representation learning lends itself to a variety of diverse models \cite{Hamilton_Ying_Leskovec_2017}. Recent advances in the literature have convincingly demonstrated the high capability of Graph Neural Networks (GNNs) \cite{Kipf_Welling_2017,Velickovic_Cucurull_Casanova_Romero_Li_Bengio_2018,Scarselli_Gori_Tsoi_Hagenbuchner_Monfardini_2009} on graph representation learning and inference. GNNs typically follow a neighborhood aggregation scheme, \ie, a node is represented by recursively aggregating other nodes' information within the neighborhood \cite{Xu_Li_Tian_Sonobe_Kawarabayashi_Jegelka_2018,Gilmer_Schoenholz_Riley_Vinyals_Dahl_2017}. Since messages pass along the edges, GNNs can effectively capture the high-order structural correlations between nodes.

Most of existing GNNs are proposed and evaluated with an underlying assumption, \ie, the distribution of collected training data is consistent with the distribution of testing data. Under this assumption, many studies follow the convention of performing random splits on graph benchmarks \cite{Kipf_Welling_2017,Xu_Hu_Leskovec_Jegelka_2019} to train and test GNNs. Unfortunately, this assumption can hardly be satisfied in real-world application-specific systems \cite{Hu_Fey_Zitnik_Dong_Ren_Liu_Catasta_Leskovec_2020,lohr2009sampling} due to the ubiquitous selection bias, leading to distribution shift between training and test data. Moreover, the test distribution is always agnostic during optimizing GNNs on training graph data. The agnostic distribution shift includes shift caused by both node labels and node attributes. For example, in recommender systems, a notorious problem is that the logged interactions collected from current undergoing systems may warp (graph) recommenders biased towards popular items, \ie, label-related distribution shift. Such a shift may degrade the recommendation effectiveness on deployed environment, \eg, causing unfairness towards less exposed items. In addition, male users often have less logged interactions on e-commerce platforms since they often online-shop with specific purposes. Therefore, the (graph) recommenders may be trained with bias towards female users and thus deteriorates the experience of male users, \ie, node attribute related distribution shift.
Generally, the agnostic distribution shift between training and testing would render traditional GNNs over-optimized on the labeled training samples and render their predictions error-prone on test samples, resulting in unstable predictions.
Therefore, learning GNNs that are resilient to distribution shifts and able to make stable predictions on graphs will be important for real-world applications.

Many techniques \cite{Bickel_Brckner_Scheffer_2009,Dudk_Schapire_Phillips_2005,Huang_Smola_Gretton_Borgwardt_Schlkopf_2006,Liu_Ziebart_2014,shimodaira2000improving} designed for general machine learning problems are proposed in the literature of stable learning. These methods typically propose to re-weight training samples with a density ratio, thus driving the training distribution closer to the testing distribution. However, how to learn structural node representations based on which we can make stable prediction on graphs across agnostic environments remains largely unexplored in the literature of graph neural networks. More recently, \cite{He_Cui_Ma_Zou_Wang_Yang_Yu_2020} proposes to learn a stable graph that captures general relational patterns, and directly optimizes the adjacency matrix with a pre-defined set generation task. However, they mainly aim to obtain an optimized and static adjacency matrix for a given graph and cannot dynamically process unseen neighborhood structures. Their framework also has the disadvantage of being tightly coupled with the set generation problem and cannot be easily adapted to other graph tasks.

In this paper, we aim to learn task-agnostic GNNs that can make stable predictions on unseen neighborhood structures. Stable prediction poses a quantitative goal that, given \textbf{\textit{one}} observational graph environment, a stably trained GNN should achieve a high average score and a low variance of scores on \textbf{\textit{multiple}} agnostic testing environments. We propose a novel stable prediction framework for GNNs, which permits both locally stable learning at the node-level, and globally stable learning at the environment-level. The proposed framework first performs biased selection on the observational environment and constructs multiple training environments. For \textit{locally stable learning}, we start from the perspective that each node is partially represented by other nodes in the neighborhood in GNNs, and we propose to capture stable properties by re-weighting the neighborhood aggregation process. Under the invariant prediction assumption for causal inference \cite{peters2016causal}, we can improve the stability of predictions when the model relies more on stable properties. For \textit{globally stable learning}, we inspect the training of GNNs at the environment-level and empirically find that the losses for different environments progressively diverge in biased training, which eventually leads to unstable performance across environments. Therefore, we propose to reduce the training gap between environments to explicitly warp the GNNs to generalize well across environments.

We conduct experiments on various public graph benchmarks and a real-world recommendation dataset that is collected from a world-leading e-commerce platform during an annual product-promotion festival where selection biases naturally exist \footnote{We will release the desensitized version to promote further investigations.}. We evaluate a bunch of generic SOTA GNNs and methods that are specifically designed for mitigating selection biases. We concern both traditional task-specific evaluation metrics and protocols that are especially designed for stable learning \cite{Kuang_Cui_Athey_Xiong_Li_2018,Kuang_Xiong_Cui_Athey_Li_2020}. Extensive results demonstrate the capability of our framework on learning GNNs that make stable predictions on graphs. In summary, the contribution of this paper are:

\begin{itemize}
	\item We propose to achieve stable prediction on graphs for explicitly reducing the performance variances across environments with distribution shifts. This is less explored in the literature.
	\item We devise a novel stable prediction framework for GNNs, which captures stable properties for each node, based on which we learn node representations and make predictions (\textit{locally stable}), and regularizes the training of GNNs on heterogeneous environments (\textit{globally stable}).
	\item We conduct comprehensive experiments on different public graph benchmarks and a noisy industrial recommendation dataset, which jointly demonstrate the effectiveness of the proposed framework.
\end{itemize}

\section{Related Works}

The recent advances in graph neural networks \cite{Wu_Souza_Zhang_Fifty_Yu_Weinberger_2019,Klicpera_Bojchevski_Gnnemann_2019,Velickovic_Cucurull_Casanova_Romero_Li_Bengio_2018,Kipf_Welling_2017,Wang_Zhu_Bo_Cui_Shi_Pei_2020,Dou_Liu_Sun_Deng_Peng_Yu_2020,Zhao_Wei_Yao_2020,Pan_Cai_Chen_Chen_Rijke_2020} have convincingly demonstrated high capability in capturing structural and relational patterns within graphs. Typically, GNNs follow a message passing schema for representation learning by transforming and aggregating the information from other nodes within the neighborhood. Different neighborhood aggregation variants \cite{Scarselli_Gori_Tsoi_Hagenbuchner_Monfardini_2009,Ding_Tang_Zhang_2018,Duan_Xiao_2019,Ma_Ahmed_Willke_Sengupta_Cole_TurkBrowne_Yu_2019,Wu_Pan_Du_Tsang_Zhu_Du_2019,Xie_Xiong_Yu_Zhu_2019,Ye_Wang_Yao_Jia_Zhou_Xiao_Yang_2019,Wang_Pan_Long_Zhu_Jiang_2017,Li_Rijke_Liu_Mao_Ma_Zhang_Ma_2020,Yang_He_Qin_Xiao_Wang_2015,Zhang_Xiong_Zhang_Zhang_Jiao_Zhu_2020,Jiao_Xiong_Zhang_Zhu_2019,Tang_Yao_Sun_Wang_Tang_Aggarwal_Mitra_Wang_2020,Zhao_Tang_Zhang_Wang_2020,Fan_Zeng_Ding_Chen_Sun_Liu_2019,Zhu_Huang_Choi_Xu_2020} have been proposed to reach state-of-the-art performances in various tasks. Typically, GAT \cite{Velickovic_Cucurull_Casanova_Romero_Li_Bengio_2018} introduces the attention mechanism into the information aggregation process. SGC \cite{Wu_Souza_Zhang_Fifty_Yu_Weinberger_2019} simplifies the original Graph Convolutional Network \cite{Kipf_Welling_2017} by linearly propagating information and collapsing weights among graph layers. APPNP \cite{Klicpera_Bojchevski_Gnnemann_2019} extends the utilized neighborhood for node representation and achieves an adjustable neighborhood for classification.

Many existing GNNs are evaluated on graph benchmarks that are randomly split \cite{Kipf_Welling_2017,Xu_Hu_Leskovec_Jegelka_2019}, \ie, the training and testing data share similar data distribution. However, in real-world applications, \eg, recommender systems, training samples are observed and collected with selection bias, leading to inconsistencies between the training and testing distribution. Few works in the literature investigate such a real-world problem. Recently, GNM \cite{NEURIPS2019_42547f5a} confronts a related problem named non-ignorable non-response, which indicates that the unlabeled nodes are missing not at random (MNAR). However, they only consider distribution caused by node labels and neglect distribution shift related to node attributes, and they solely discuss binary-class datasets in the experiments, which can be less plausible for real-world scenarios. In this paper, we propose a framework that can alleviate the negative effects from shift related to both labels and attributes, and obtains stable predictions on various graph benchmarks and real-world recommendation datasets. \cite{He_Cui_Ma_Zou_Wang_Yang_Yu_2020} proposes to learn a static adjacency matrix for a given graph and expects that the learned adjacency matrix captures general relational patterns that are free from selection biases. We largely differ from this work as illustrated in the Introduction.

Recently, many methods \cite{Bickel_Brckner_Scheffer_2009,Dudk_Schapire_Phillips_2005,Huang_Smola_Gretton_Borgwardt_Schlkopf_2006,Liu_Ziebart_2014,shimodaira2000improving} are proposed to address selection bias for general machine learning problems. They typically align the training distribution with the testing distribution by re-weighting the training samples. Following these works, \cite{Kuang_Cui_Athey_Xiong_Li_2018,Kuang_Xiong_Cui_Athey_Li_2020} first propose the stable learning framework as well as the evaluation protocols. They mainly decorrelate different dimensions of the hidden variable, and they prove that the prediction based on decorrelated variables should be stable. More recently, GAT-DVD \cite{biasdebiased} directly borrows such a decorrelation idea to the literature of GNNs. In this paper, instead of borrowing generic off-the-shelf stable learning tools, we propose to inspect the neighborhood aggregating process and design graph-specific architectures. Besides, we show a large improvement over GAT-DVD in the experiments.

\section{Problem Formulation} \label{sec:probfor}

Let $\mathcal{X}$ denote the space of observed features, $\mathcal{A}$ denote the space of adjacency matrix, and $\mathcal{Y}$ denote the outcome space. Following \cite{peters2016causal}, we define a graph \textbf{environment} as the joint distribution $P_{XAY}$ on $\mathcal{X} \times \mathcal{A} \times \mathcal{Y}$ and use $\mathcal{E}$ to denote the set of all environments. For each environment, we have a graph dataset $G^e = (\mathbf{X}^e, \mathbf{A}^e, Y^e)$, where $\mathbf{X}^e \in \mathcal{X}$ are node features, $\mathbf{A}^e \in \mathcal{A}$ is the adjacency matrix, and $Y^e \in \mathcal{Y}$ is the response variable (\eg, node labels in the node classification problem). The joint distribution of features and outcomes on $(\mathbf{X}, \mathbf{A}, Y)$ can vary across environments, \ie, $P_{X A Y}^{e} \neq P_{X A Y}^{e^{\prime}}$ for $e, e^{\prime} \in \mathcal{E}$, and $e \neq e^{\prime}$. In this paper, we aim to learn node representations based on which we can make stable predictions across environments with various degrees of selection biases.  Before giving the specific definition of stable prediction, we first define $Average\_Score$ and $Stability\_Score$ similar to \cite{Kuang_Cui_Athey_Xiong_Li_2018} of a GNN model as:
\begin{align}
	Average\_Score &= \frac{1}{|\mathcal{E}|} \sum_{e \in \mathcal{E}} \operatorname{S}\left(G^{e}\right) \label{eq:avgscore}, \\
	Stability\_Error &= \sqrt{\frac{1}{|\mathcal{E}|-1} \sum_{e \in \mathcal{E}}\left(\operatorname{S}\left(G^{e}\right)-Average\_Score\right)^{2}}. \label{eq:stability_error}
\end{align}
where $|\mathcal{E}|$ denotes the number of environments evaluated, and $\operatorname{S}(G^e)$ refers to the predictive score computed with a specific numerical assessment on dataset $G^e$. Therefore, $Average\_Score$ indicates the overall performance of the learned GNN on heterogeneous testing environments, and $Stability\_Error$ indicates how differently the learned GNN performs on these environments. Based on these two metrics, we define the problem of stable prediction on graphs:

\begin{problem}[Stable Prediction on Graphs] 
\textbf{Given} one training environment $e\in \mathcal{E}$ with dataset $G^{e}=(\mathbf{X}^{e}, \mathbf{A}^e, Y^{e})$, the task is to \textbf{learn} node representations based on which the predictions yield high $Average\_Score$ but small $Stability\_Error$ across environments $\mathcal{E}$.
\end{problem}

\section{Methods}

\begin{figure*}[t] \begin{center}
    \includegraphics[width=\textwidth]{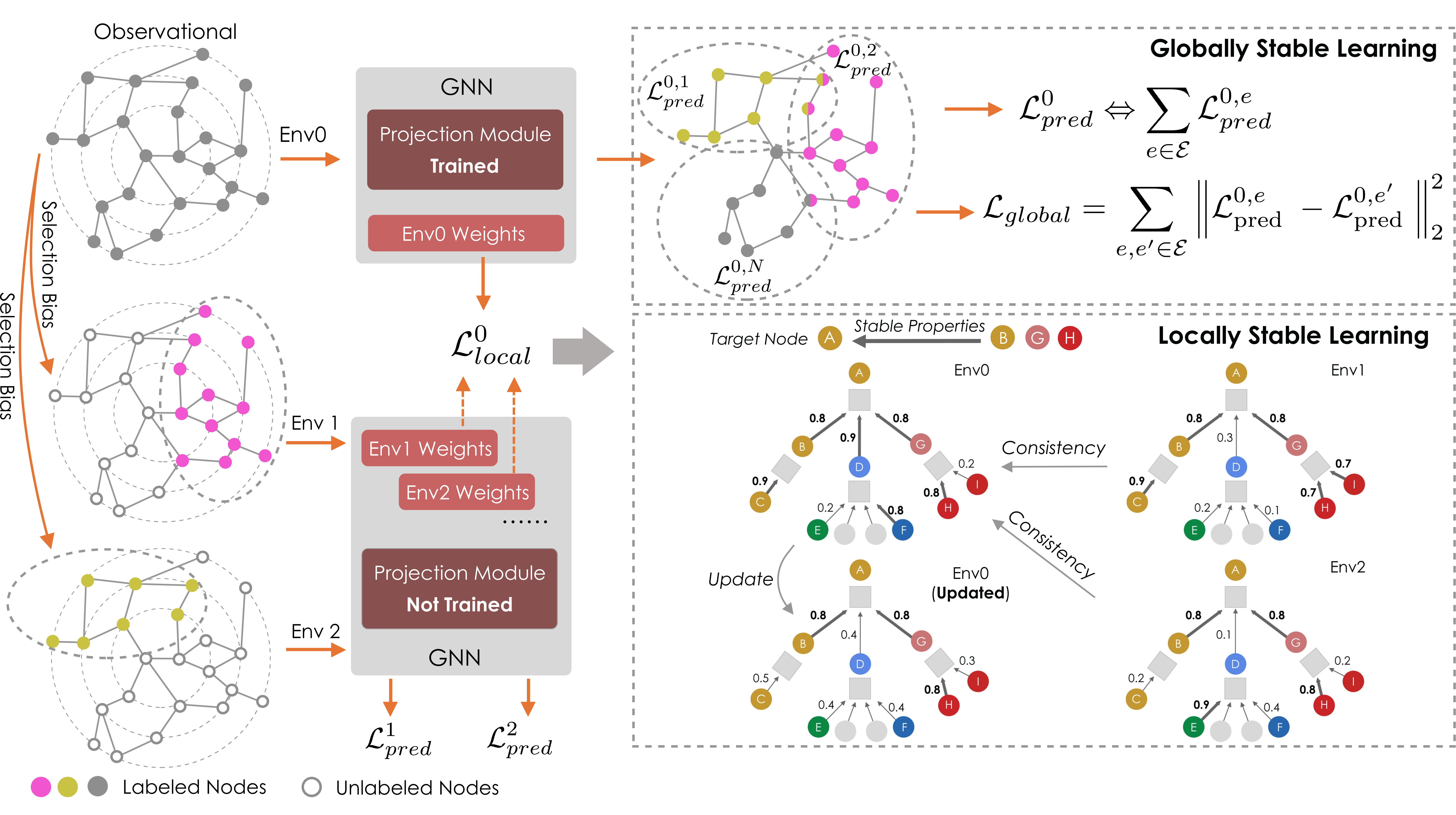}
    \caption{
    The overall schema of the proposed framework, which consists of two essential components, \ie, the \textit{locally stable learning} that captures properties that are stable across environments in the representation learning of each target node, and the \textit{globally stable learning} that explicitly balances the training of different environments.
	}
\label{fig:schema}
\end{center} \end{figure*} 

\subsection{A Retrospect of Modern GNNs}

Here we briefly summarize typical GNNs and then give an illustration on why they suffer from the distribution shift. Modern GNNs follow a neighborhood aggregation schema by iteratively aggregating representations of its neighbors to update the representation of the target node. One iteration can be generally formulated as:
\begin{align}
	\mathbf{x}^{*}_i &= \operatorname{ AGGREGATE } \left(\left\{ \mathbf{x}_j : j \in \mathcal{N}_{i}\right\}\right), \\ 
	\mathbf{x}^{\prime}_i &= \operatorname{COMBINE} \left(  \mathbf{x}^{*}_i, \mathbf{x}_i  \right).
\end{align}
where $\mathcal{N}_i$ denotes the indices of nodes in the neighborhood of node $i$, $\mathbf{x}^{\prime}_i$ denotes the updated representation of node $i$. The modeling of $\operatorname{ AGGREGATE }$ and $\operatorname{COMBINE}$ can be various and essential for different GNN architectures. Most GNNs are optimized to obtain an overall best performance on the observational environment, \ie, $G^0 = (\mathbf{X}^0, \mathbf{A}^0, Y^0)$. In this way, we may expect an overly-optimized solution after training and some distribution-specific patterns may warp the GNN biased towards a globally sub-optimal solution since the distribution of real-world testing data can be agnostic and mostly yields shift from the training graph distribution in real-world applications (\eg, recommender systems). We denote this phenomenon as \textit{global instability}.

In the local view, \ie, neighborhood aggregation process, each neighbor node contributing to the final aggregated representation can be viewed as a property of the root node. Most GNNs are proposed disregarding a fundamental problem, \ie, "what are the \textbf{stable} properties of the root node?" The connection between capturing stable properties and stable prediction can be interpreted as a plausible hypothesis:
\begin{hypothesis}
	If we consider all stable properties ("direct causes") of the target node of interest, then the conditional distribution of the outcome given the stable properties will not change when interventions (\eg, biased selection) are performed without affecting the target and stable properties themselves.
\end{hypothesis}
This stability idea is closely linked to causality and has been discussed or empirically demonstrated to be effective in the literature \cite{haavelmo1944probability,aldrich1989autonomy,pearl2009causality}. In other words, we can achieve stable prediction on graphs across environments with various selection biases by capturing the stable properties for each node. Indeed, some existing GNNs (\eg, GAT \cite{Velickovic_Cucurull_Casanova_Romero_Li_Bengio_2018}) propose to specify different weights to different nodes in a neighborhood and somewhat "stabilize" the weights learning process by leveraging multiple heads. However, we argue that the weights they learned explore the subtle statistical relationship in training data and can be far from being stable due to potential confounders \cite{pearl2016causal}. A confounder denotes the common cause of the predictor and the outcome, which may lead to spurious correlations between the root node (outcome) and some "important" neighbors (predictor). We denote this phenomenon as \textit{local instability}.

\subsection{Stable Learning on Graphs}

The essence of our methodology for stable learning on graphs is depicted in Figure \ref{fig:schema}. Given the environments priorly generated by various selection biases, we propose the \textit{locally stable learning} strategy, which explicitly captures the stable properties in learning the representation of each node by re-weighting the information propagation/aggregation processes, and the \textit{globally stable learning}, which regularizes the errors across environments to be closer and thus improving the stability. 
We borrow the idea from the literature of causality \cite{peters2016causal} and propose to first perform several interventions on the observational data based on selection biases, and create some environments $\mathcal{E}$ accordingly. We note that we keep an observational environment in $\mathcal{E}$.


\subsubsection{Locally Stable Learning} \label{sec:localstable}

In the neighborhood aggregation process, we regard nodes that are consistently important across environments as stable properties, as depicted in Figure \ref{fig:schema}. Therefore, the final representation of the target node of interest is primarily based on such stable properties. 

Given one environment $e$, we obey a weighted schema of the commonly used neighborhood aggregation in most GNNs:
\begin{align}
	\mathbf{x}_{i}^{\prime}=\sigma\left(\sum_{j \in \mathcal{N}_{i}} \alpha_{i j}^e \mathbf{W} \mathbf{x}_{j}\right),
\end{align}
where $\mathbf{x}_{i}^{\prime}$ denotes the updated representation of the target node $x_{i}$, $\mathbf{W}$ denotes the weight matrix of the linear transformation, $\sigma$ denotes the nonlinear activation function, $\mathcal{N}_{i}$ denotes the indices set of node $x_{i}$'s neighbors, and $\alpha_{i j}^e$ is the weight that indicates the importance of one property $x_{j}$ to the target node $x_{i}$ learned in environment $e$. The weight $\alpha_{i j}^e$ can be learned in various ways and we adopt the formulation of graph attention as the weight predictor $\varphi^e$ for its simplicity:
\begin{align}
	\alpha^e_{i j}=\frac{\exp \left(\text { LeakyReLU }\left(\mathbf{a}^{e}\left[\mathbf{W} \mathbf{x}_{i} \| \mathbf{W} \mathbf{x}_{j}\right]\right)\right)}{\sum_{k \in \mathcal{N}_{i}} \exp \left(\text { LeakyReLU }\left(\mathbf{a}^{e}\left[\mathbf{W} \mathbf{x}_{i} \| \mathbf{W} \mathbf{x}_{k}\right]\right)\right)}
\end{align}
where LeakyReLU denotes the nonlinearity (with negative input slope 0.2), and $\mathbf{a}^{e}$ denotes the parameter vector to determine the weight $\alpha^e_{i j}$ for environment $e$.
The weights in graph attention indicate relative importance in the neighborhood, and it is also acceptable to use the following formulation, which has a sense of absolute importance:
\begin{align}
		\alpha^e_{i j}=\text { sigmoid }\left(\mathbf{a}^{e}\left[\mathbf{W} \mathbf{x}_{i} \| \mathbf{W} \mathbf{x}_{j}\right]\right), \label{eq:recngcf}
\end{align}
We note that each environment $e$ has its parameter vector $\mathbf{a}^{e}$, which helps find important properties especially for this environment (or data distribution). To identify and leverage properties that are consistently important across all environments $\mathcal{E}$, a straightforward solution is to treat properties with weights higher than a certain threshold as stable properties and manually encourage the model to rely on them for prediction while suppressing the others. However, this solution has at least two drawbacks: 1) manually modifying the weights of properties will lead to inconsistencies between the current weight prediction module and other modules in GNN since they will be trained based on the modified weights rather than the predicted weights; 2) some properties that are not important may happen to obtain high weights during training and model regarding them as stable properties may draw false conclusions. 

In this regard, we propose to identify and leverage stable properties in a soft way: 1) we propose to pull the weight predicted across environments by using a distance loss; 2) we will regard properties with weights that are both high in value and can be easily pulled together as stable properties and suppress the others. In this way, we can confront the randomness of gradient descent based training (\ie, the second problem) by pulling sensitive weights towards the corresponding insensitive ones, and confront the inconsistent training problem (\ie, the first problem) since other parts in GNNs will always reply on the currently predicted weights. To further simplify the training process, we note that the first pulling process can also help to suppress properties that yield inconsistent weights since they will be driven towards the average. Therefore, the locally stable regularizer can be written as the following:
\begin{align}
	\mathcal{L}_{local}^{e} = \sum_{i \in \mathcal{V}} \sum_{j \in \mathcal{N}_i}  \sum_{e' \in \mathcal{E}} \operatorname{Dist}(\alpha^e_{i j}, \alpha^{e'}_{i j}), \label{eq:local}
\end{align}
where $\mathcal{V}$ denotes the indices set of all nodes, and $\operatorname{Dist}$ denotes a distance function which is set as the L2 distance in our experiment for its empirical effectiveness. When there are multiple GNN layers equipped with the re-weighting module, the regularizer can be written as:
\begin{align}
	\mathcal{L}_{local}^{e} = \sum_{l=1}^{N_l} \mathcal{L}_{local}^{e,l}.
\end{align}
where $N_l$ denotes the number of GNN layers equipped with the re-weighting module.

\subsubsection{Globally Stable Learning}

Besides investigating stable prediction from a local view, \ie, learning node representations that capture stable properties, we further investigate stable prediction from a global and environment-level view. Recall that one of the fundamental goals of stable prediction is to reduce $Stability\_Error$, \ie, the standard deviation of scores across environments. However, although we can individually compute scores for all environments, it is not directly applicable to optimize objectives related to non-differentiable metrics. We start with a perspective that model performances correlate well with training losses and empirically find that, in the observational environment, the sub-losses that belong to different environments progressively diverge during training with selection bias. Such divergence in training losses will eventually lead to gaps in testing performance. In this regard, we propose to explicitly reduce the gap between losses across environments to rectify the unstable training. Mathematically, the proposed stable regularizer for globally stable learning can be formularized as the following:
\begin{align}
	\mathcal{L}_{global}=\sum_{e, e^{\prime} \in \mathcal{E}} \left( \mathcal{L}_{pred}^{0,e} - \mathcal{L}_{pred}^{0,e'} \right)^{2}, \label{eq:global}
\end{align}
where $\mathcal{L}_{pred}^{0}$ denotes the task-specific loss computed in the observational environment, \ie, environment 0, and $\mathcal{L}_{pred}^{0,e}$ denotes the sub-loss that belongs to environment $e$, \ie,
\begin{align}
	\mathcal{L}_{pred}^{0,e} = \sum_{l \in \mathcal{Y}_{L}^{0}} \mathbbm{1}(l \in \mathcal{Y}_{L}^{e} ) \mathcal{L}_{pred,l}^{0}. \label{eq:globalper}
\end{align} 
where $\mathcal{Y}_{L}^{e}$ denotes the indices set of nodes that have labels for environment e. We note that minimizing the pair-wise distance of losses as defined in Equation \ref{eq:global} is equivalent to minimizing the variance of losses. 
Since the negative of loss function can be interpreted as a certain scoring function, it is equivalent to minimizing the stability error defined in Equation \ref{eq:stability_error}.

\subsubsection{Training} \label{sec:training}

With the proposed two building blocks, \ie, locally and globally stable learning, we give a detailed illustration of how we train the entire framework. The training procedure is summarized in Algorithm \ref{ag:stableprediction}. Given one observational environment 0 for training with dataset $G^{0}=(\mathbf{X}^{0}, \mathbf{A}^0, Y^{0})$, we perform biased selection with one or more factors (\eg, node label or semantic node attributes) on $G^{0}$ at the beginning of training or each training epoch. After selection, we obtain several environments $\mathcal{E}$ with the corresponding graph datasets $\{G^{e}=(\mathbf{X}^{e}, \mathbf{A}^e, Y^{e})\}_{e = 0, \dots, \left|\mathcal{E}\right|}$. Each environment keeps a weight predictor that is individually trained, and all the environments share the same GNN backbone, which is solely trained on environment 0, \ie, the observational environment. We train the entire framework following an environment-by-environment procedure and update the corresponding parameters with the task-specific objective $\mathcal{L}_{pred}^{e}$ (\eg, node classification, recommendation). For the observational environment, we train the weight predictor for this environment and the GNN backbone with combined loss function:
\begin{align}
	\mathcal{L} = \mathcal{L}_{pred}^{0} + \lambda_0 \mathcal{L}_{local}^{0} + \lambda_1 \mathcal{L}_{global}. \label{eq:all}
\end{align}
We do not train other environments with the corresponding $\mathcal{L}_{local}^e$ since we mainly aim to rectify the importance weights in the observational environment. We note that the GNN backbone and the weight predictor in the observational environment will be used for testing/inference.

{\SetAlgoNoLine%

	\begin{algorithm}[!t]
		\DontPrintSemicolon
  
  \KwInput{Observational graph datasets $G^{0}=(\mathbf{X}^{0}, \mathbf{A}^0, Y^{0})$ }
  \KwOutput{Parameters of stably learned GNN backbone $\theta$ and weight predictors $\varphi^0$ for the observational environment}
	
	Perform biased selection on $G^{0}$ and obtain $\{G^{e}=(\mathbf{X}^{e}, \mathbf{A}^e$ $, Y^{e})\}_{e = 1}^{\left|\mathcal{E}\right|}$. Initialize $\theta$ and $\{ \varphi^e \}_{e=0}^{\left|\mathcal{E}\right|}$
	
	\While{not converged}
	{
		Freeze $\theta$ \tcp*{Disable gradients}
		\For{e = 1 to $\left|\mathcal{E}\right|$}
		{
			Optimize $\varphi^e$ to minimize $\mathcal{L}_{pred}^{e}$, and cache $\mathbf{\alpha}^e$
		}
		Unfreeze $\theta$ \tcp*{Enable gradients}
		Compute $\mathcal{L}_{pred}^0$ and $\mathcal{L}_{global}$ as in Eq. \ref{eq:global}-\ref{eq:globalper}, and compute $\mathcal{L}_{local}^{0}$ with cached $\{ \mathbf{\alpha}^e \}_{e=1}^{\left|\mathcal{E}\right|}$ as in Eq. \ref{eq:local}.
		
		Optimize  $\theta, \varphi^0$ to minimize $\mathcal{L}$ as in Eq. \ref{eq:all}.
	}
	
\caption{Stable Learning on Graphs}
\label{ag:stableprediction}
\end{algorithm}

}%

%

\section{Experiments}

\begin{figure*}[!t]
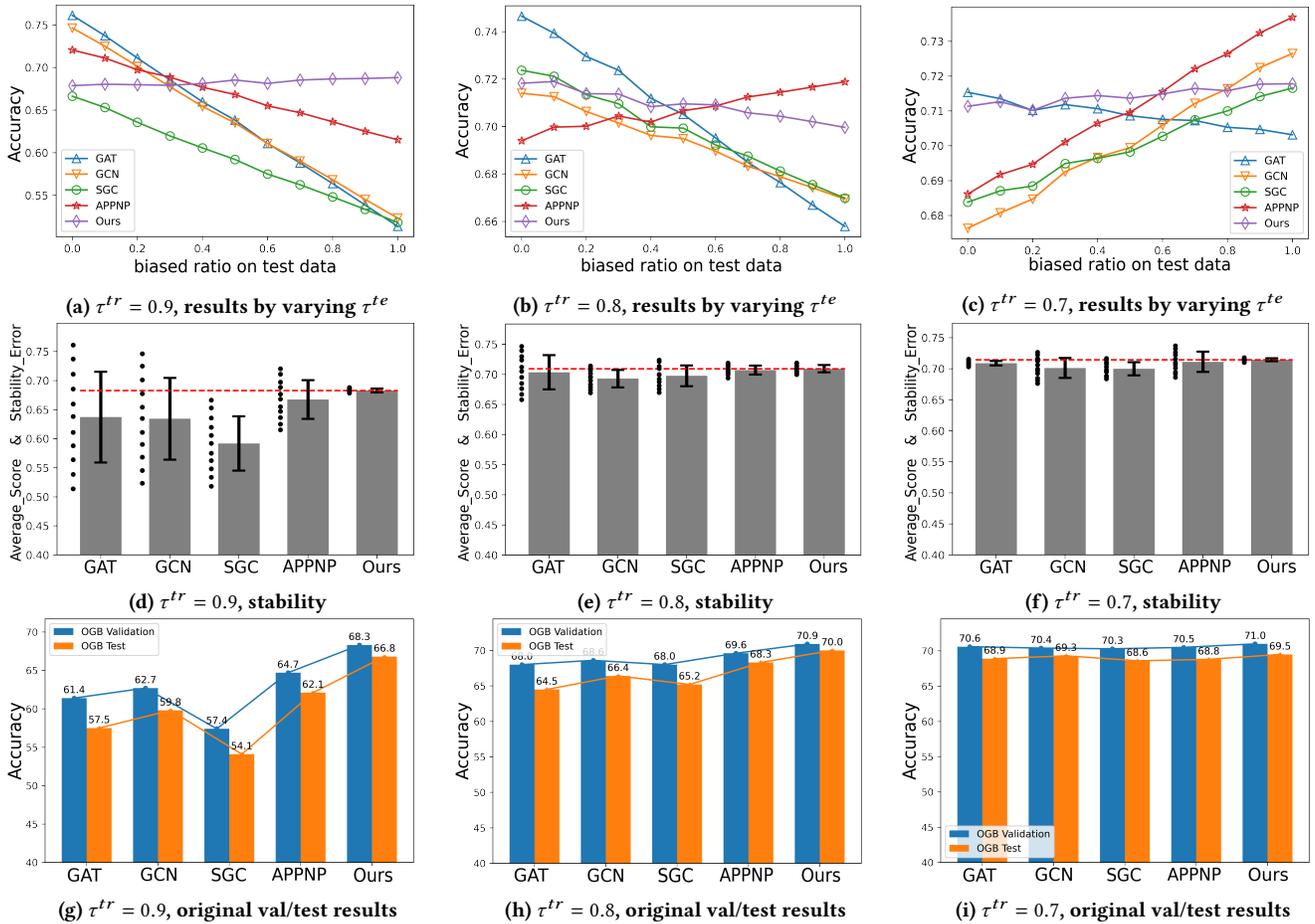
 \begin{center}
\begin{subfigure}{\subfigwidth\textwidth}
	\includegraphics[width=0.92\linewidth]{./figures/arxiv_0901_line.pdf}
    \caption{
    $\tau^{tr} = 0.9$, results by varying $\tau^{te}$
    	}
\label{fig:arxiv_0901_line}
\end{subfigure}
\begin{subfigure}{\subfigwidth\textwidth}
	\includegraphics[width=0.92\linewidth]{./figures/arxiv_0802_line.pdf}
    \caption{
    $\tau^{tr} = 0.8$, results by varying $\tau^{te}$
    	}
\label{fig:arxiv_0802_line}
\end{subfigure}
\begin{subfigure}{\subfigwidth\textwidth}
	\includegraphics[width=0.92\linewidth]{./figures/arxiv_0703_line.pdf}
    \caption{
    $\tau^{tr} = 0.7$, results by varying $\tau^{te}$
    	}
\label{fig:arxiv_0703_line}
\end{subfigure}

\begin{subfigure}{\subfigwidth\textwidth}
    \includegraphics[width=0.92\linewidth]{./figures/arxiv_0901_stability.pdf}
    \caption{
    $\tau^{tr} = 0.9$, stability
    	}
\label{fig:arxiv_0901_stability}
\end{subfigure}
\begin{subfigure}{\subfigwidth\textwidth}
    \includegraphics[width=0.92\linewidth]{./figures/arxiv_0802_stability.pdf}
    \caption{
    $\tau^{tr} = 0.8$, stability
    	}
\label{fig:arxiv_0802_stability}
\end{subfigure}
\begin{subfigure}{\subfigwidth\textwidth}
    \includegraphics[width=0.92\linewidth]{./figures/arxiv_0703_stability.pdf}
    \caption{
    $\tau^{tr} = 0.7$, stability
    	}
\label{fig:arxiv_0703_stability}
\end{subfigure}

\begin{subfigure}{\subfigwidth\textwidth}
    \includegraphics[width=0.92\linewidth]{./figures/arxiv_0901_oritest.pdf}
    \caption{
    $\tau^{tr} = 0.9$, original val/test results
    	}
\label{fig:arxiv_0901_oritest}
\end{subfigure}
\begin{subfigure}{\subfigwidth\textwidth}
    \includegraphics[width=0.92\linewidth]{./figures/arxiv_0802_oritest.pdf}
    \caption{
    $\tau^{tr} = 0.8$, original val/test results
    	}
\label{fig:arxiv_0802_oritest}
\end{subfigure}
\begin{subfigure}{\subfigwidth\textwidth}
    \includegraphics[width=0.92\linewidth]{./figures/arxiv_0703_oritest.pdf}
    \caption{
    $\tau^{tr} = 0.7$, original val/test results
    	}
\label{fig:arxiv_0703_oritest}
\end{subfigure}

    \caption{
    Testing results on OGB-Arxiv dataset by varying the training bias ratio $\tau^{tr}$ among $\{ 0.9, 0.8, 0.7 \}$. Subfigures \ref{fig:arxiv_0901_line} - \ref{fig:arxiv_0703_line} show the metrics by varying the testing bias ratio. Subfigures \ref{fig:arxiv_0901_stability} - \ref{fig:arxiv_0703_stability} explictly show the stability of prediction by plotting the $Average\_Score$ and $Stability\_Error$. Subfigures \ref{fig:arxiv_0901_oritest} - \ref{fig:arxiv_0703_oritest} show the results on the original validation and testing datasets.
    	}
    \label{fig:ogbarxiv}
\end{center} \end{figure*}

We note that instability in prediction arises due to the inconsistencies of the joint distribution of features, adjacency matrix, and outcomes $(\mathbf{X}, \mathbf{A}, Y)$ across environments (\eg, training/testing environments) according to Section \ref{sec:probfor}. To test the stability of all methods, for each dataset, we construct several training and testing environments by varying the joint distribution of $(\mathbf{X}, Y)$. We consider both shift related to node labels by directly varying $Y$ and shift related to node attributes by varying additional factors that well correlate with $\mathbf{X}$. We disregard the shift related to $\mathbf{A}$ in the experiment.

Following previous GNNs \cite{Velickovic_Cucurull_Casanova_Romero_Li_Bengio_2018,Kipf_Welling_2017} , we evaluate the proposed stable learning framework on public graph benchmarks, including the recently proposed Open Graph Benchmark \cite{Hu_Fey_Zitnik_Dong_Ren_Liu_Catasta_Leskovec_2020}, in Section \ref{sec:bench}. Since these datasets seldom accompany meaningful attributes, we are mainly concerned with \textit{label}-related shift. We conduct hyper-parameter analysis to obtain a better understanding of the designs. We further demonstrate its efficacy on real-world recommendation scenarios that are known to be full of sample selection biases \cite{Chen_Dong_Wang_Feng_Wang_He_2020} in Section \ref{sec:rec}. We mainly report results with shift caused by node attributes as a complement to label shift on graph benchmarks. We also consider the settings with agnostic real-world selection biases in Section \ref{sec:agnostic}.

\subsection{Experiments on Graph Benchmarks} \label{sec:bench}

\subsubsection{Experimental Setup} We conduct experiments on the following datasets, of which the statistics are listed in Table \ref{tab:datadesc}:

\begin{table}[!t]
    \caption{Statistics of datasets.}
    \vspace{-0.1in}
    
    \begin{subtable}{1\columnwidth}
    \caption{Dataset statistics.}
    \centering
    \small
    \begin{tabular}{l|rrr}
        \toprule
        Dataset & \textbf{Citeseer} &  \textbf{OGB-Arxiv} & \textbf{Recommendation}  \\\midrule
          Nodes & 3,327  &   169,343  & 29,444   \\
          Edges & 4,732  & 1,166,243   & 180,792       \\\bottomrule
    \end{tabular}
    \label{tab:SLdatadesc}
    
    \end{subtable}

%

    \begin{subtable}{1\columnwidth}
    \caption{Detailed statistics of the recommendation dataset, which contains 5 consecutive days collected during a product promotion festival. (\textit{IteracN} stands for the number of Interactions at Day N.) }
    \centering
    \small
    \begin{tabular}{ll  rrrrr}
        \toprule
          \textbf{Users} & \textbf{Items} & \textbf{Iterac1} & \textbf{Iterac2} & \textbf{Iterac3}  & \textbf{Iterac4}  & \textbf{Iterac5}    \\\midrule
          9,052  & 20,392 &   180,792  & 161,035 &  174,511  & 181,166  &  233,373 \\\bottomrule
    \end{tabular}
    \label{tab:privatedatadesc}
    \end{subtable}

    \label{tab:datadesc}
    \normalsize
\end{table}


\vpara{OGB Arxiv.} The OGB datasets are proposed with real-world (nonrandom) training/testing split, which is suitable for stability evaluation.  We perform biased selection on the half nodes from the original training set based on their corresponding labels (\ie, label shift) to construct the observational environment. We construct multiple testing environments from the remaining half nodes. For each environment, the probability of node $i$ with label $y$ to be selected can be $P\left(s_{i}=1 \right) = \tau$ if $y \geq 24$ and $1-\tau$ otherwise. We note that other selection choices are acceptable and we here mainly aim to get originally equal-size environments by splitting in the middle, \ie, 24.
$\tau$ denotes the bias ratio indicating the severity of sample selection bias. We use $\tau^{tr}, \tau^{te}$ to denote the training and testing bias ratio, respectively. In the experiments, we vary the training bias ratio $\tau^{tr}$ among $\{ 0.9, 0.8, 0.7\}$ indicating heavy, medium, and light selection biases, following \cite{Kuang_Cui_Athey_Xiong_Li_2018}.

\begin{figure*}[!t] \begin{center}
\vspace{0.2cm}
\begin{subfigure}{.33\textwidth}
	\includegraphics[width=0.93\linewidth]{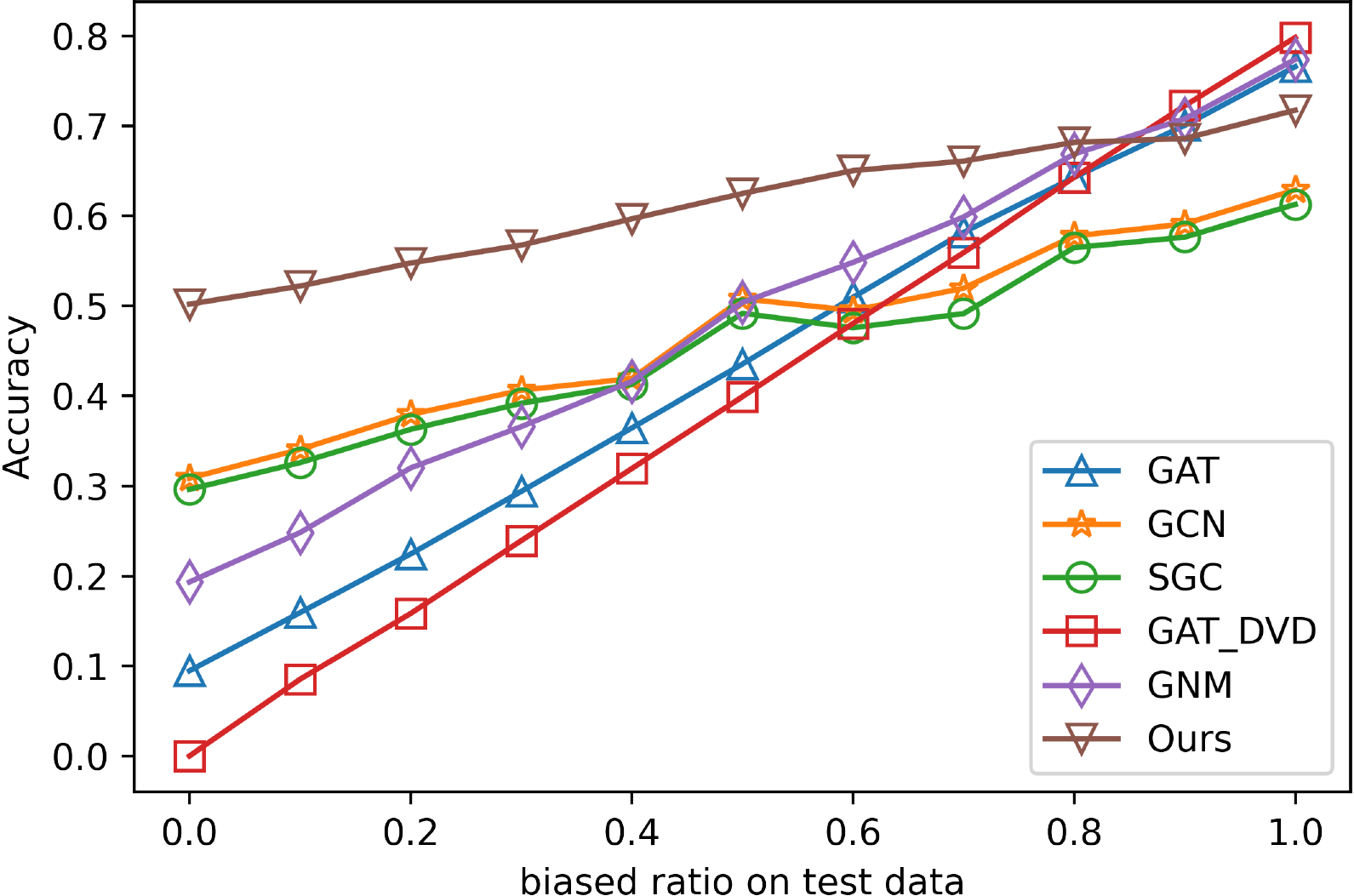}
    \caption{
    $\tau^{tr} = 0.8$, results by varying $\tau^{te}$
    	}
\label{fig:citeseer_0802_line}
\end{subfigure}
\begin{subfigure}{.33\textwidth}
	\includegraphics[width=0.93\linewidth]{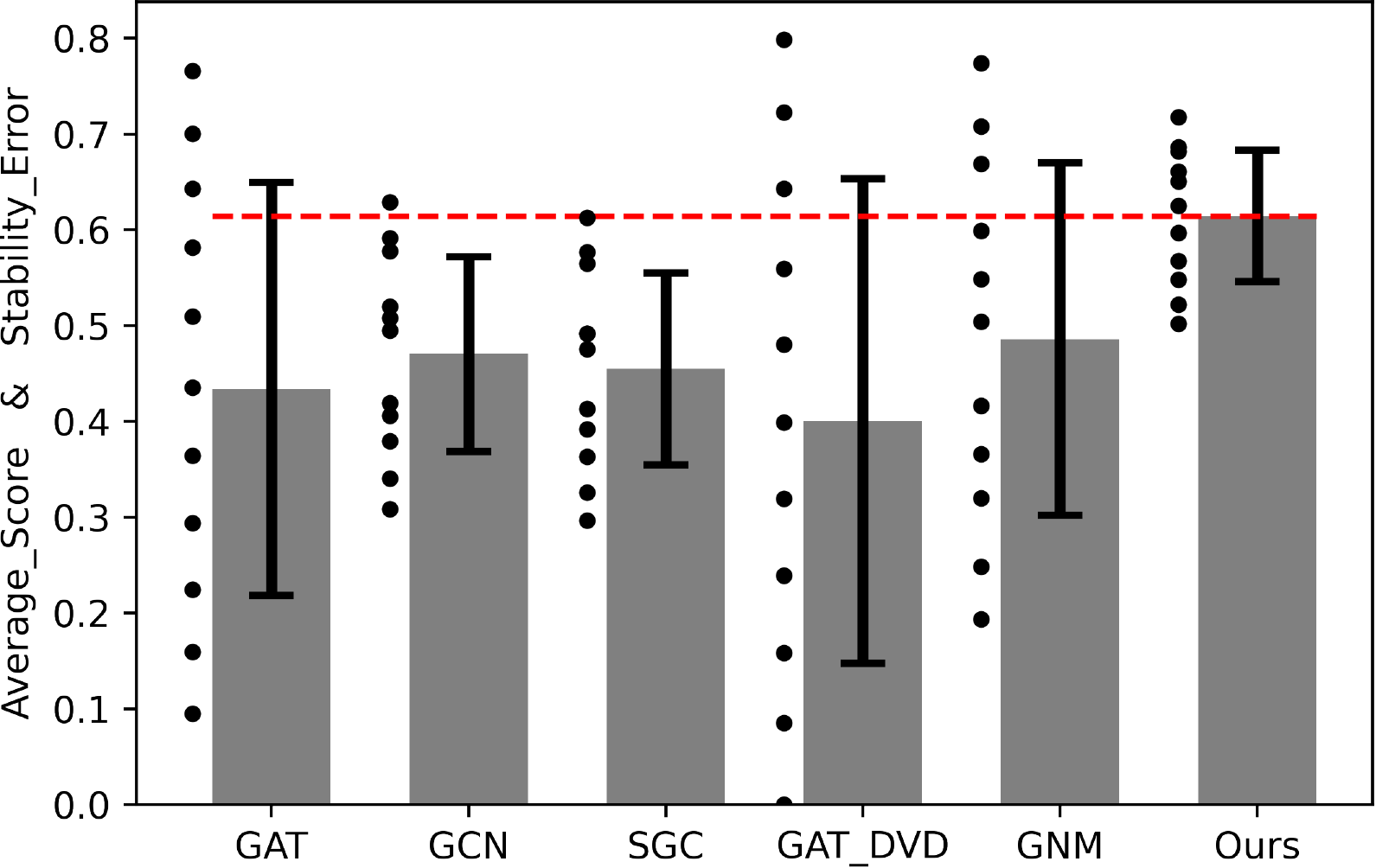}
    \caption{
    $\tau^{tr} = 0.8$, stability
    	}
\label{fig:citeseer_0802_stability}
\end{subfigure}
\begin{subfigure}{.33\textwidth}
	\includegraphics[width=0.93\linewidth]{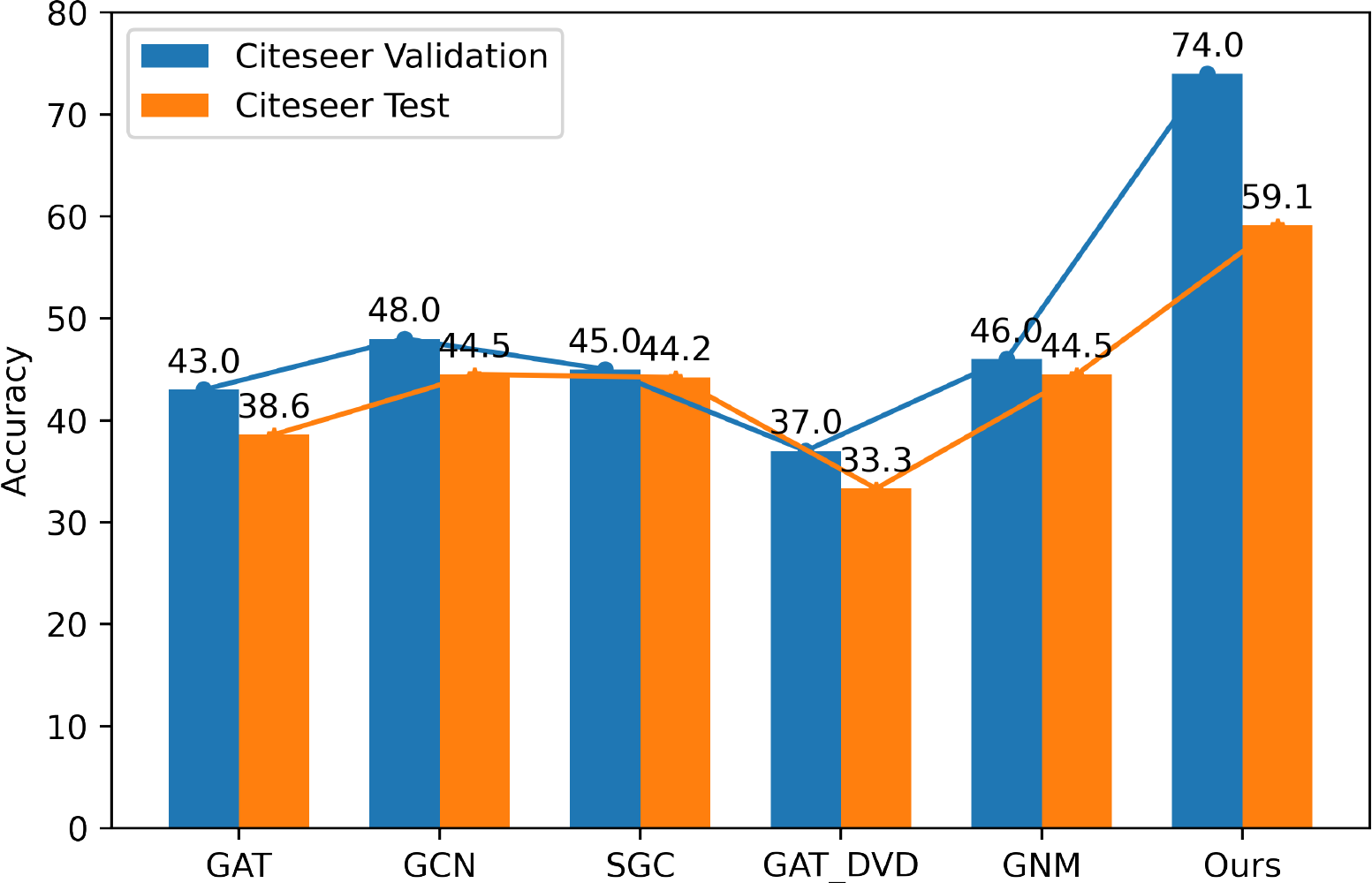}
    \caption{
    $\tau^{tr} = 0.8$, original val/test results
    	}
\label{fig:citeseer_0802_orivaltest}
\end{subfigure}

    \caption{
    Testing results on Citeseer dataset.
    	}
    \label{fig:citeseer}
\end{center} \end{figure*}

We consider several SOTA GNNs as comparison methods, including GCN \cite{Kipf_Welling_2017}, GAT \cite{Velickovic_Cucurull_Casanova_Romero_Li_Bengio_2018}, SGC \cite{Wu_Souza_Zhang_Fifty_Yu_Weinberger_2019}, and APPNP \cite{Bojchevski_Klicpera_Perozzi_Kapoor_Blais_Rzemberczki_Lukasik_Gnnemann_2020}.

\vpara{Citeseer.} We also report the results on a traditional benchmark, Citeseer \cite{Sen_Namata_Bilgic_Getoor_Gallagher_Eliassi_Rad_2008}. The probability of node $i$ with label $y$ to be selected can be $P\left(s_{i}=1 \right) = \tau$ if $y \geq 3$ and $1-\tau$ otherwise.
We further add two comparison methods that are designed for alleviating selection biases, \ie, GNM \cite{NEURIPS2019_42547f5a}, and GAT-DVD \cite{biasdebiased}. We do not test them on OGB-Arxiv since they are not easily adaptable for large-scale datasets. We consider training bias ratio $\tau^{tr}=0.8$, which is the same as the medium-level selection bias in the Arxiv dataset.

We follow \cite{Kipf_Welling_2017,Velickovic_Cucurull_Casanova_Romero_Li_Bengio_2018} to construct two-layer GCN and GAT. All other methods (including ours) also contain two graph layers for a fair comparison. All methods are with hidden size 250 for OGB-Arxiv and 64 for Citeseer, and learning rate 0.002.


%
%
%
%


\subsubsection{Stability comparison with SOTA GNNs.} The testing results on the OGB-Arxiv dataset and Citeseer dataset are shown in Figure \ref{fig:ogbarxiv} and Figure \ref{fig:citeseer}. Specifically, each figure in Figure \ref{fig:arxiv_0901_line} - \ref{fig:arxiv_0703_line} plot the evaluation results across different testing environments with bias ratio $\tau^{te}$ varying among $\{ 0.0, 0.1, 0.2, \dots, 1.0 \}$. Figure \ref{fig:arxiv_0901_line} - \ref{fig:arxiv_0703_line} differ from each by the training bias ratio $\tau^{tr}$. Since stable prediction expects both low performance variances across multiple testing environments and an overall high performance, we explicitly plot the corresponding Stability\_Error and Average\_Score in Figure \ref{fig:arxiv_0901_stability} - \ref{fig:arxiv_0703_stability}. In a nutshell, our framework achieves more stable results than SOTA GNNs, including both generic ones (GAT, GCN, SGC, and APPNP) and those designed for reducing selection biases (GAT-DVD and GNM). In different testing environments constructed by varying the testing bias ratio, we observe that most GNNs suffer from the distributional shifts and yield poorer performances when the testing distribution is more different from the training distribution (\eg, the right of Figure \ref{fig:arxiv_0901_line}). Although our framework sacrifices some performance in testing environments with distribution closer to the training (\eg, the left of Figure \ref{fig:arxiv_0901_line}), our framework obtains significantly higher Average\_Score and lower Stability\_Error, which are important metrics (as illustrated in Section \ref{sec:probfor}) for stable prediction \cite{Kuang_Cui_Athey_Xiong_Li_2018}, across heterogeneous environments as indicated in Figure \ref{fig:arxiv_0901_stability} - \ref{fig:arxiv_0703_stability}, \ref{fig:citeseer_0802_stability}.

By varying the training bias ratio from \textit{light} to \textit{heavy} (from Figure \ref{fig:arxiv_0703_stability} to \ref{fig:arxiv_0901_stability}), we can observe that the stability errors are increasing significantly for most conventional GNNs. This means that they are sensitive to the selection biases and the resulted distribution shifts. When the distributional shifts are heavier, the performance across multiple testing environments yields larger differences, \ie, unstable predictions. Our framework yields the least stability error that can be almost negligible, which demonstrates the superiority. On the other hand, when the distribution shifts are increasing from \textit{light} to \textit{heavy}, the least Average\_Score drop of our framework demonstrates that we are not achieving some trivial solutions. For example, one trivial solution is to deteriorate the performance on data distributions similar to the training environments, which can somehow reduce the Stability\_Error but have the cost of deteriorating the overall effectiveness. In a nutshell, our framework achieves stable prediction on graphs without sacrificing effectiveness. Although APPNP yields high Average\_Score compared to other SOTA GNNs, their performance scores across different testing environments mostly vary significantly (by taking an in-depth analysis on Figure \ref{fig:arxiv_0703_line} or \ref{fig:arxiv_0901_line}), \ie, unstable predictions.



Surprisingly, with the same hyper-parameters, GAT-DVD and GNM achieve more unstable results than other generic GNNs. As shown in Figure \ref{fig:citeseer_0802_stability}, GAT-DVD and GNM yield similar or better Average\_Score but significantly larger Stability\_Error compared to other SOTA GNNs. We attribute these results to that GNM is designed for binary-class datasets and may perform poorly or need further improvements when extending it to multi-classes datasets, and that the work of GAT-DVD \cite{biasdebiased} is still under development.

\subsubsection{Performance on the original validation/testing datasets.} Since the Open Graph Benchmark splits the datasets under real-world settings, we also report the performance on the original OGB validation and test datasets, which can be viewed as real-world environments. As shown in Figure \ref{fig:arxiv_0901_oritest}-\ref{fig:arxiv_0703_oritest}, our framework consistently outperforms the other methods. The improvement can be larger when the bias is more severe. These results again verifies the importance of alleviating selection biases and the effectiveness of our framework.

\begin{figure}[!t] \begin{center}
\begin{subfigure}{.23\textwidth}
	\includegraphics[width=0.99\linewidth]{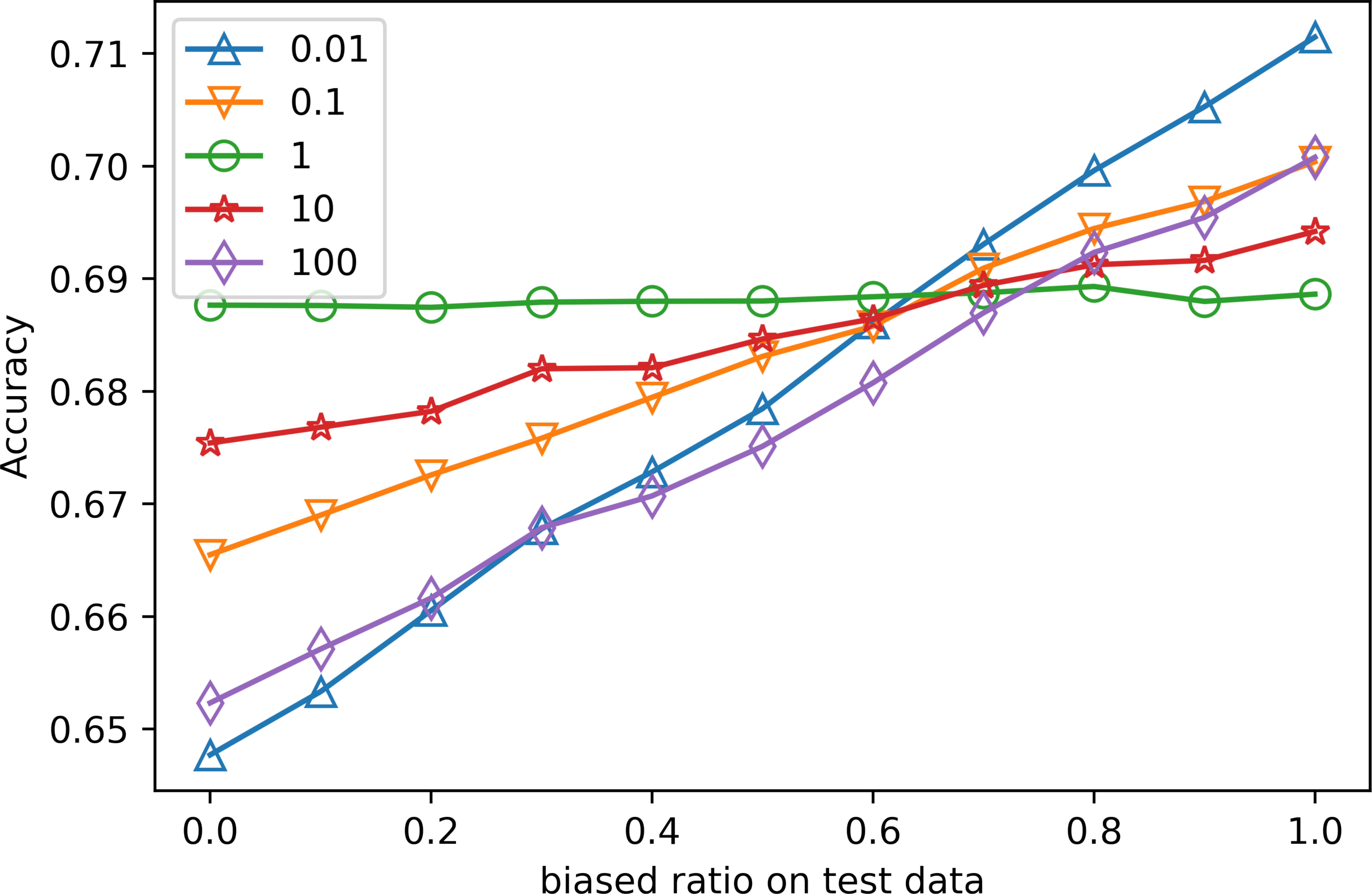}
    \caption{
    	\footnotesize{Analysis on $\lambda_0$}, varying $\tau^{te}$
    	}
\label{fig:hypermi_line}
\end{subfigure}
\begin{subfigure}{.23\textwidth}
	\includegraphics[width=0.99\linewidth]{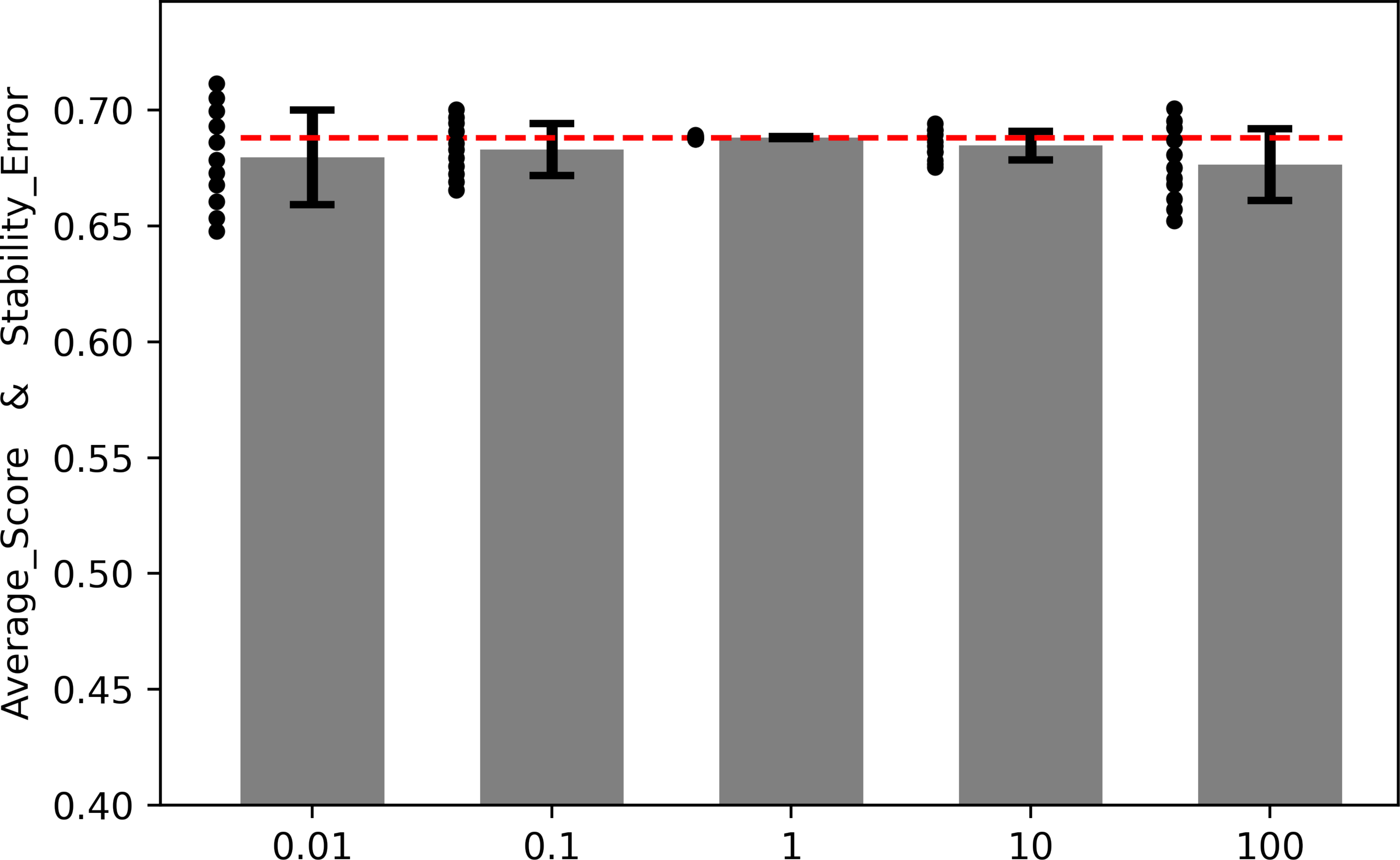}
    \caption{
    \footnotesize{Analysis on $\lambda_0$}, statbility
    	}
\label{fig:hypermi_stability}
\end{subfigure}

\begin{subfigure}{.23\textwidth}
    \includegraphics[width=0.99\linewidth]{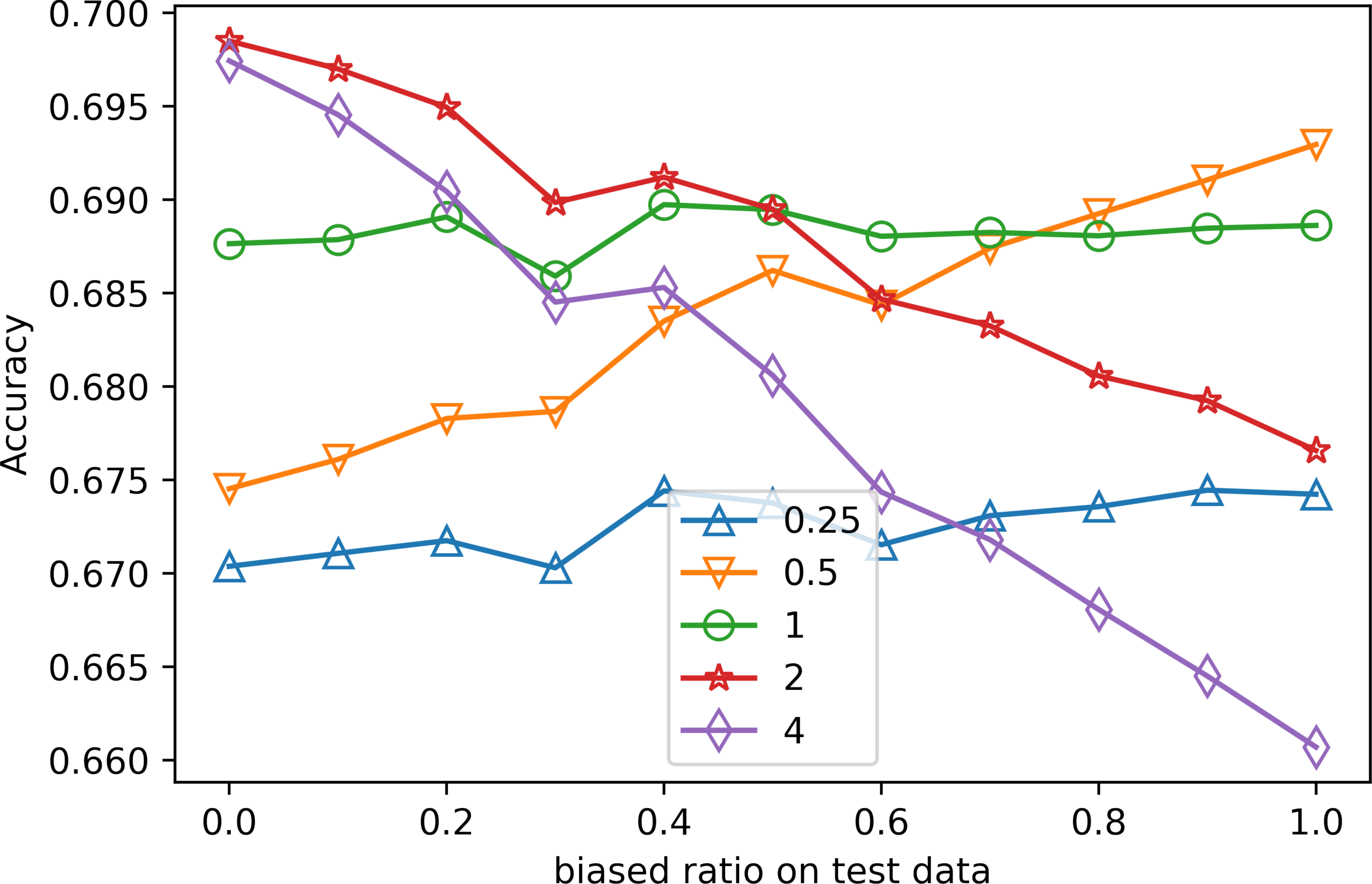}
    \caption{
    \footnotesize{Analysis on $\lambda_1$}, varying $\tau^{te}$
    	}
\label{fig:hypersr_line}
\end{subfigure}
\begin{subfigure}{.23\textwidth}
    \includegraphics[width=0.99\linewidth]{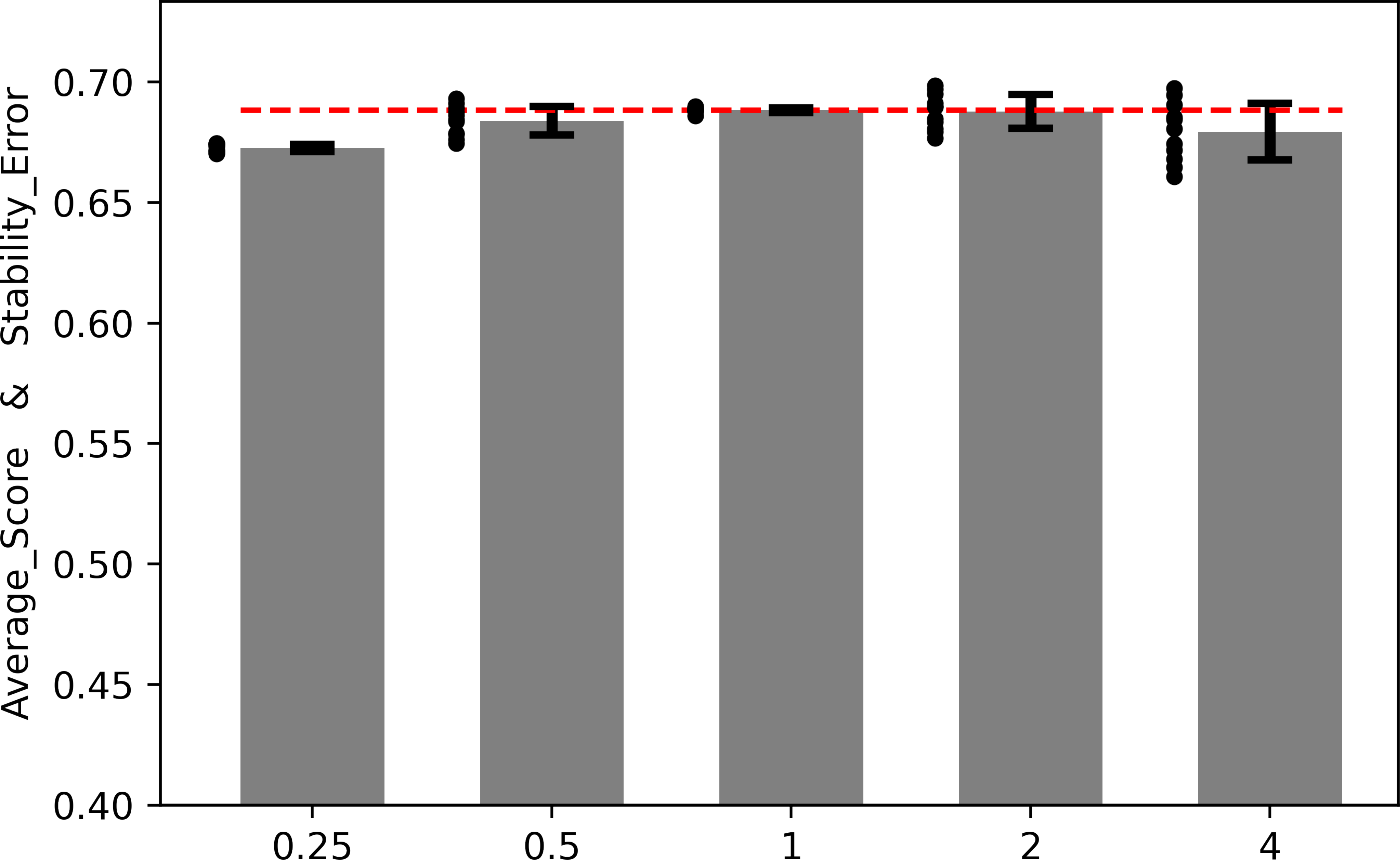}
    \caption{
    \footnotesize{Analysis on $\lambda_1$}, statbility
    	}
\label{fig:hypersr_stability}
\end{subfigure}

    \caption{
    Hyper-parameter analysis on the OGB-Arxiv dataset with training bias ratio $\tau^{tr}=0.8$.
    	}
    	
    \label{fig:hyper}
\end{center} \end{figure}


\subsubsection{Analysis on Hyper-parameters} To investigate how the proposed two regularizers affect the stability of prediction, we vary the corresponding hyper-parameters, \ie, $\lambda_0$ and $\lambda_1$, on the OGB-Arxiv dataset with training bias ratio $\tau^{tr}=0.8$. Figure \ref{fig:hyper} summarizes the results. When increasing the $\lambda_0$ and $\lambda_1$ from a lower rate, we observe a significant stability improvement, \ie, improved Average\_Score and reduced Stability\_Error. This demonstrates the effectiveness of our framework in the sense of ablation study. In other words, progressively reducing the impact of the two regularizers to a lower rate will lead to a performance drop. We also observe that when the impact of the two regularizers becomes increasingly larger, \eg, $\lambda_0 = 10$ or $100, \lambda_1 = 2$ or $4$, there is a performance drop in the model's stability. We attribute this phenomenon to that a large $\lambda$ of one regularizer (global or local) may deteriorate the capability of the other regularizer (local or global) as well as the task-specific capacity, thus leading to unstable predictions. When comparing the global and local regularizers, the effectiveness of local stable learning is less sensitive to the corresponding hyper-parameter $\lambda_0$. In other words, we can easily improve the model stability using the local regularizer without cautiously tuning $\lambda_0$. As for the global regularizer, relatively small or large $\lambda_1$ will harm the stability. This suggests that universally pulling the losses of each node across heterogeneous environments at the $same$ rate might not be an optimal solution. Therefore, we leave finding adaptive pulling strategies for different nodes and environments as a promising future work.

\subsection{Experiments on a Recommendation Dataset} \label{sec:rec}

\subsubsection{Experimental Setup} To demonstrate the efficacy of the proposed framework on real-world applications, we conduct experiments on the user-item bipartite graph of recommender systems, where sample selection bias are arguably ubiquitous \cite{Chen_Dong_Wang_Feng_Wang_He_2020}. 

\vpara{Collecting a Real-world \textit{Noisy} Dataset.} Specifically, we collect an industrial dataset from one of the world-leading e-commerce platforms from June 11th, 2020, to June 15th, 2020, when an annual product promotion festival is being celebrated. In such a period, the promotion strategies from online shop owners can be various and time-evolving. Therefore, inconsistencies between the users' clicks and their satisfactions naturally exist, leading to a distribution shift from the collected data and the real-world testing environment. We select users and items that have interactions in all five days, which means we do not consider the cold-start setting. We mainly consider click interactions, which is a common setting for the deep candidate generation phase in recommendation. We further select users that have 200-300 interactions to ensure the quality of data. We split the interactions into several environments according to the day they happen, and the statistics are listed in Table \ref{tab:datadesc}. We use data samples from the first day for training and use the remaining for evaluation. We keep the gender and age attributes for users. For users that do not provide these attributes, we use the value predicted by the e-commerce platform. There are 8 age sections provided by the platform, including 1-18, 19-25, 26-30, 31-35, 36-40, 41-50, 51-60, and >=61. We group 1-18 and 19-25 into a holistic section and the remaining into another section. This split results in approximately equally-sized sections.

\vpara{Evaluation} We mainly focus on the deep candidate generation stage of recommendation. The user-item bipartite graph consists of user nodes and item nodes. A user is connected to an item when the user interacts with the item. Solely id feature is considered and transformed to dense vectors using a learnable embedding matrix, which is a common practice for deep candidate generation models \cite{He_Deng_Wang_Li_Zhang_Wang_2020,Wang_He_Wang_Feng_Chua_2019}. The embedding matrix is learned along with the graph recommenders. The collected dataset contains rich semantic attributes, and we discuss user gender/age in experiments, respectively. The probability of a user to be selected can be $P\left(s_{i}=1 \right) = \tau$ if $age \leq 25\ or\ gender\ =\ M$ and $1-\tau$ if $age > 25\ or\ gender\ =\ F$.
$M$ denotes male and $F$ denotes female. We also discuss agnostic selection bias by viewing each of the following days as an individual environment. We incorporate a widely used metric NDCG \cite{Wang_He_Wang_Feng_Chua_2019,He_Deng_Wang_Li_Zhang_Wang_2020}. Compared to other widely used metrics such as Recall and Hit Ratio, NDCG considers the positions of recommended items. NDCG can be formally written as:
	\begin{align}
		\text{DCG@N} &= \frac{1}{|\mathcal{U}|} \sum_{u \in \mathcal{U}} \sum_{r \in \hat{\mathcal{I}}_{u, N}} \frac{\mathbbm{1}(r \in \mathcal{I}_{u})}{\log _{2}\left(i_{r}+1\right)} \\
		\text{NDCG@N} &= \frac{ \text{DCG@N} }{\text{IDCG@N}}
	\end{align}
where $\mathcal{U}$ is the set of users, N is the number of recommended items, and $\hat{i}_{u, k}$ indicates the $k$th item recommended for user $u$. $\mathbbm{1}$ denotes the indicator function. $\text{IDCG@N}$ denotes the ideal discounted cumulative gain and is the maximum possible value of $\text{DCG@N}$. consider the top 100 generated candidates of each model for evaluation, \ie, NDCG@100. We report the percentage score in the results.

\vpara{Baselines} We consider the following models as baselines:

\begin{itemize}[leftmargin=*]
	\item \textit{NGCG} leverages GCN to represent users and items based on the user-item bipartite graph.
	\item \textit{LightGCN} linearly propagates user/item embeddings on the user-item graph and thus simplifying the NGCF.
	\item \textit{Stable Graph Recommender.} We build the proposed stable graph recommender based on NGCF. NGCF largely follows the standard GCN model. The proposed stable graph recommender propagates embeddings on the user-item bipartite graph as the following:
\begin{align}
&\alpha^e_{u i}=\text { sigmoid }\left(\mathbf{a}^{e}\left[\mathbf{e}_{u}^{0} \| \mathbf{e}_{i}^{0}\right]\right) \\
&\mathbf{e}_{u}^{(k+1)}=\sigma\left(\mathbf{W}_{1} \mathbf{e}_{u}^{(k)}+\sum_{i \in \mathcal{N}_{u}} \frac{\alpha^e_{u i}}{\sqrt{\left|\mathcal{N}_{u}\right|\left|\mathcal{N}_{i}\right|}}\left(\mathbf{W}_{1} \mathbf{e}_{i}^{(k)}+\mathbf{W}_{2}\left(\mathbf{e}_{i}^{(k)} \odot \mathbf{e}_{u}^{(k)}\right)\right)\right) \\
&\mathbf{e}_{i}^{(k+1)}=\sigma\left(\mathbf{W}_{1} \mathbf{e}_{i}^{(k)}+\sum_{u \in \mathcal{N}_{i}} \frac{\alpha^e_{u i}}{\sqrt{\left|\mathcal{N}_{u}\right|\left|\mathcal{N}_{i}\right|}}\left(\mathbf{W}_{1} \mathbf{e}_{u}^{(k)}+\mathbf{W}_{2}\left(\mathbf{e}_{u}^{(k)} \odot \mathbf{e}_{i}^{(k)}\right)\right)\right)
\end{align}
where $\mathbf{e}_{u}^{(k+1)}$ and $\mathbf{e}_{i}^{(k+1)}$ denote the updated user and item embedding after $k$ layers propagation. $\sigma$ denotes a certain nonlinear activation function and is $\operatorname{LeakyReLU}$ as NGCF. $\mathcal{N}_{u}$ denotes the set of interacted items for user $u$ and $\mathcal{N}_{i}$ denotes the set of interacted users for item $i$. $\mathbf{W}_{1}$ and $\mathbf{W}_{2}$ are learnable transformation matrices. $\alpha^e_{u i}$ denotes the importance of interaction $<u,i>$ for representing user $u$ and item $i$. We note that interactions that are consistently important across environments $\mathcal{E}$ are stable properties, as illustrated in Section \ref{sec:localstable}. We keep a global $\alpha^e_{u i}$ for layers due to its efficiency and do not compute weights per layer. We note that deep candidate generation models, which recall Top K items from a billion-scale item gallery are largely sensitive to model efficiency. For example, one of the contribution of LightGCN is to remove the nonlinearity $\sigma$ in NGCF and thus improving efficiency. We train the stable graph recommender as illustrated in Section \ref{sec:training}.
\end{itemize}


\begin{figure}[!t] \begin{center}
\begin{subfigure}{.235\textwidth}
	\includegraphics[width=1.1\linewidth]{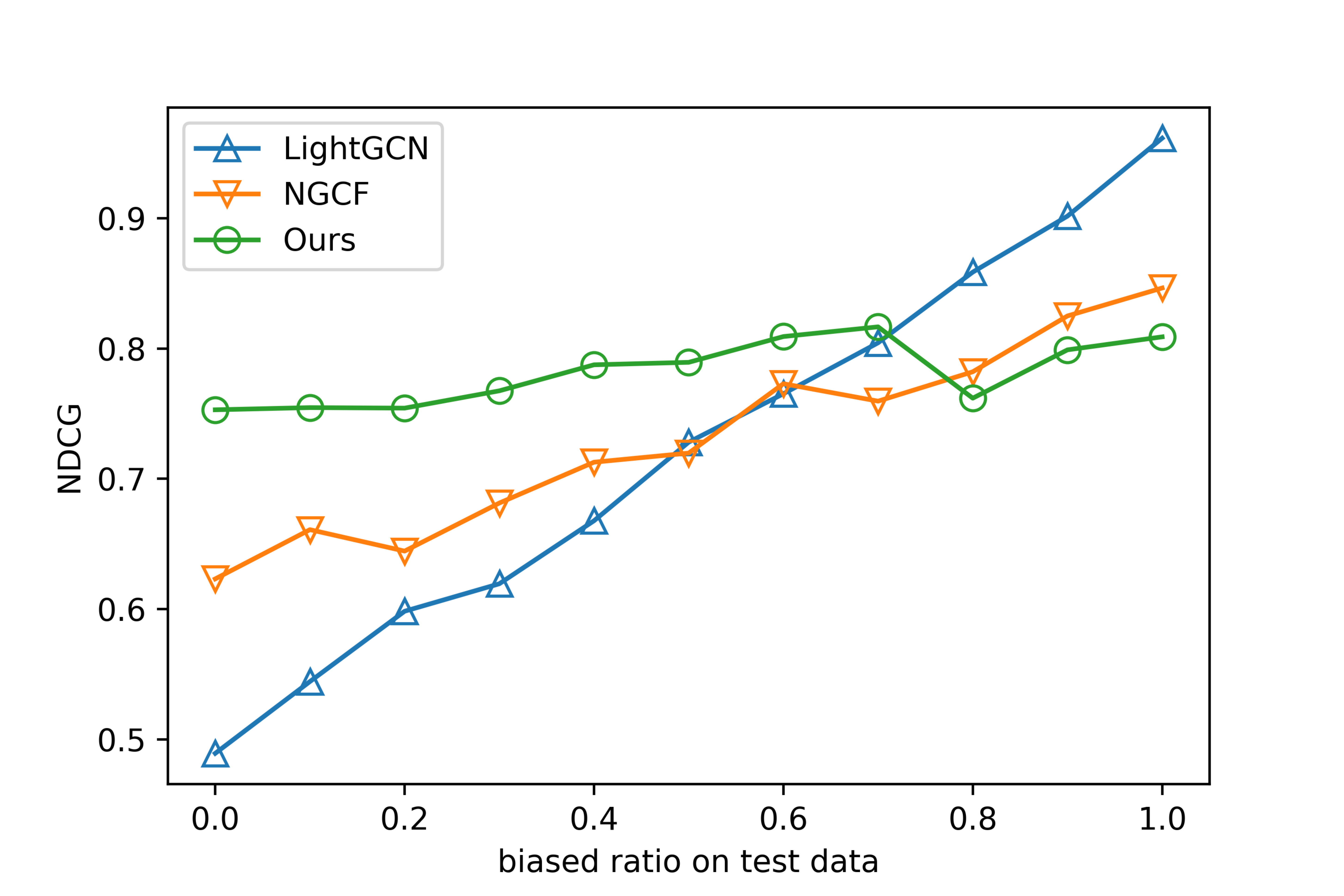}
    \caption{
   \footnotesize{Age, $\tau^{tr}=0.6$}, varying $\tau^{te}$
    	}
\label{fig:taobao618_age_0604_line}
\end{subfigure}
\begin{subfigure}{.235\textwidth}
	\includegraphics[width=1.1\linewidth]{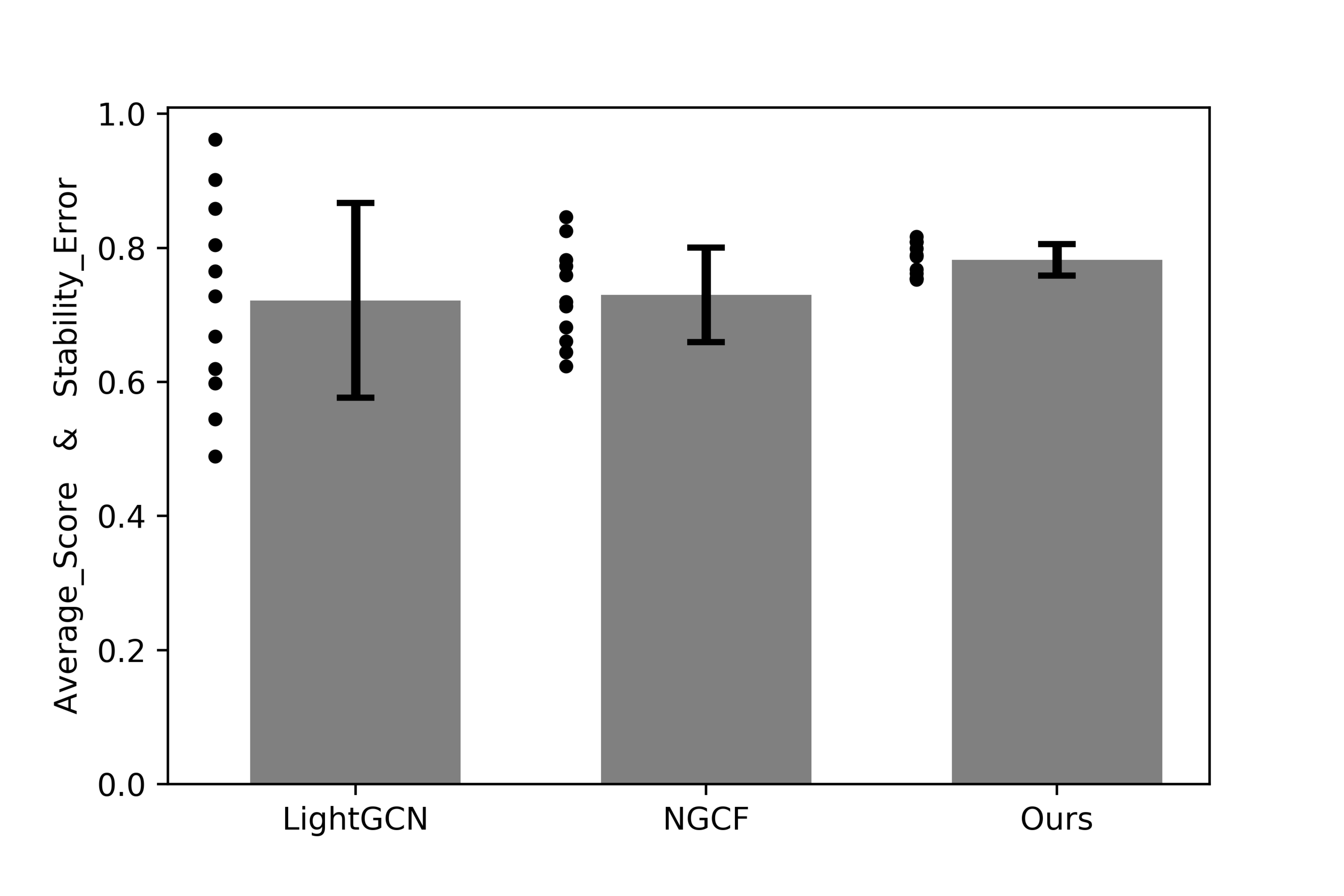}
    \caption{
    \footnotesize{Age, $\tau^{tr}=0.6$}, stability
    	}
\label{fig:taobao618_age_0604_stability}
\end{subfigure}

\begin{subfigure}{.235\textwidth}
    \includegraphics[width=1.1\linewidth]{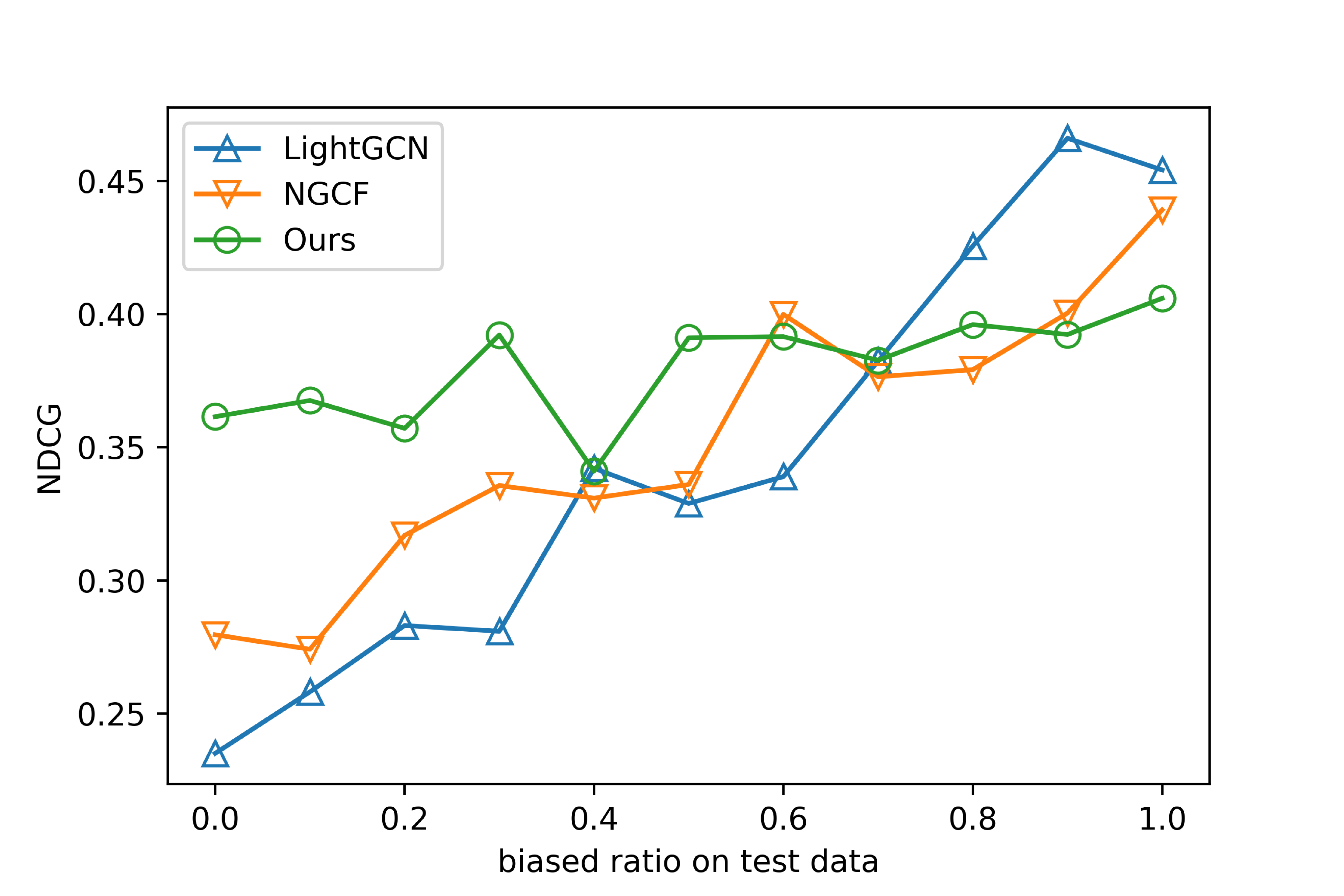}
    \caption{
   \footnotesize{Gender, $\tau^{tr}=0.6$}, varying $\tau^{te}$
    	}
\label{fig:taobao618_gender_0604_line}
\end{subfigure}
\begin{subfigure}{.235\textwidth}
    \includegraphics[width=1.1\linewidth]{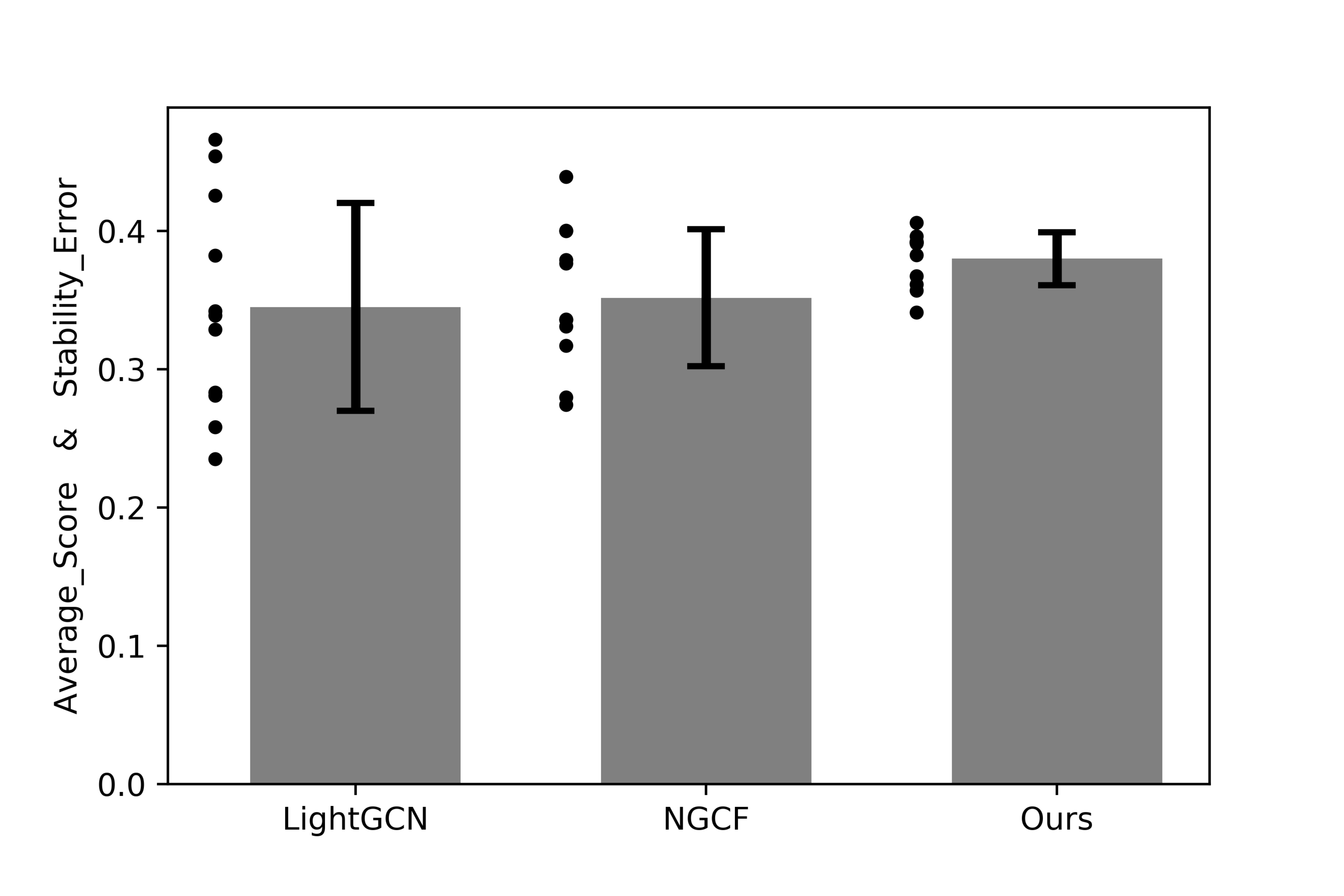}
    \caption{
    \footnotesize{Gender, $\tau^{tr}=0.6$}, stability
    	}
\label{fig:taobao618_gender_0604_stability}
\end{subfigure}

    \caption{
    Results on real-world recommendation dataset with distribution shift caused by node attributes.
    	}
    \label{fig:rec}
\end{center} \end{figure}

\begin{figure}[!t] \begin{center}
\begin{subfigure}{.235\textwidth}
	\includegraphics[width=1.1\linewidth]{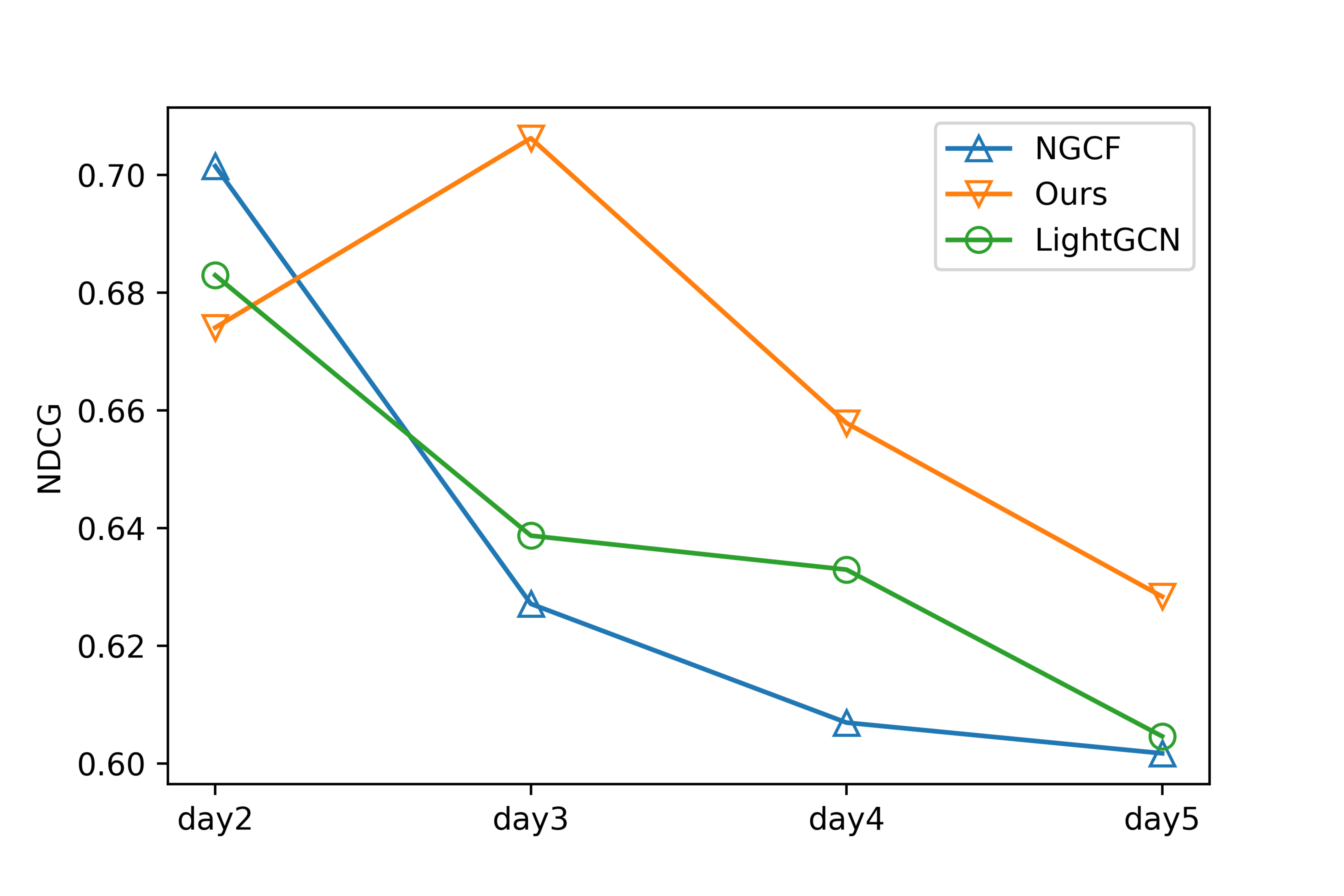}
    \caption{
    \footnotesize{Age, $\tau^{tr}=0.6$}
    	}
\label{fig:taobao618_age_0604_stability}
\end{subfigure}
\begin{subfigure}{.235\textwidth}
	\includegraphics[width=1.1\linewidth]{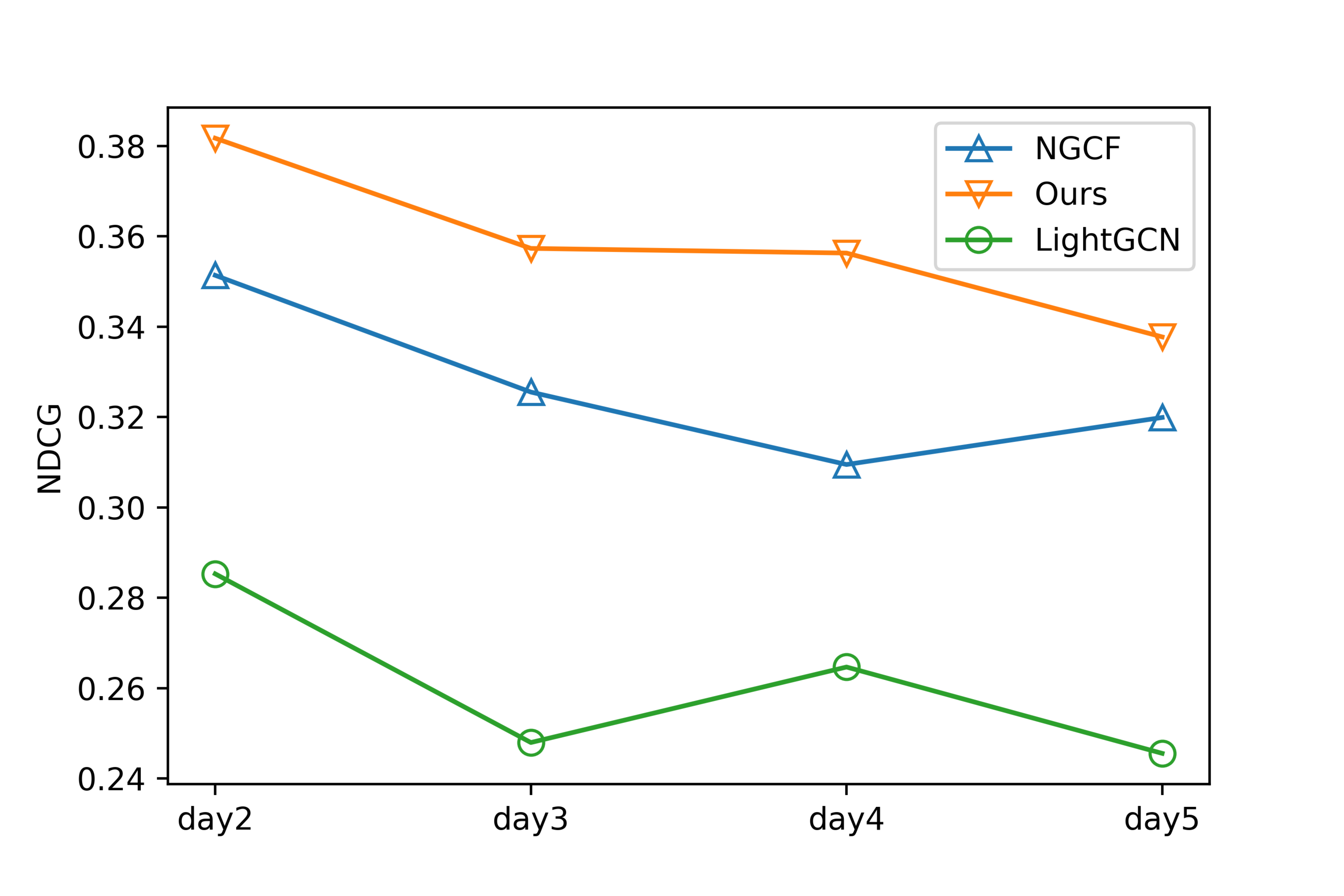}
    \caption{
   \footnotesize{Gender, $\tau^{tr}=0.6$}
    	}
\label{fig:taobao618_age_0604_line}
\end{subfigure}

    \caption{
    Results on recommendation dataset with real-world environments (each day as an individual environment).
    	}
    \label{fig:agnostic}
\end{center} \end{figure}

%
%
%

\subsubsection{Analysis on Stable Prediction} The testing results with distribution shift caused by node attributes are shown in Figure \ref{fig:rec}. Specifically, we set training bias rate $\tau^{tr}=0.6$ and test the model on testing environments with bias rate varying among $\{ 0.0, 0.1, 0.2, \dots, 1.0 \}$. Overall, we observe that our framework achieves significantly more stable results than the other two SOTA graph recommenders. Specifically, the Average\_Score is significantly improved with Stability\_Error largely reduced. Noteworthy, similar to the findings to those on the graph benchmarks, the stable graph recommender improves the stability mostly by boosting the performance on environments where NGCG/LightGCN performs poorly. For example, as shown in Figure \ref{fig:taobao618_age_0604_line}, NGCF/LightGCN achieves about 0.6/0.5 NDCG score when $\tau^{tr}=0.6$ and $\tau_{te}=0.0$ while the proposed stable graph recommender significantly boost the NDCG to nearly 0.77 with the same distribution shift. These results further demonstrate the effectiveness of the proposed method on broader real-world applications.


\subsubsection{Evaluation with Real-world Environments} \label{sec:agnostic}

To evaluate our framework on real-world environments with agnostic selection bias, we construct testing environments from days that follow the day where graph recommenders are trained. To be specific, the model is trained on the recommendation data logged on June 11th, 2020, and we test the trained model on the data logged on June 12-15th, 2020, individually. Therefore, there are four testing environments in total, and we report the results in Figure \ref{fig:agnostic}. We observe that there is a consistent performance improvement across heterogeneous real-world environments. These results suggest that our framework is capable of capturing stable properties that should persist across time in node representation learning. For example, a user might click some items due to external distractions, which can hardly represent the user's inherent interest. This phenomenon is known as the \textit{natural noise} \cite{Mahony_Hurley_Silvestre_2006}, and such clicks are unstable properties when representing the user node on the user-item bipartite graph. The proposed stable graph recommender captures non-noisy interactions for users on the user-item bipartite graph and obtains stable/inherent interest representations that are effective across time.


\section{Conclusion and Future Work}

In this paper, we argue that real-world applications (\eg, recommender systems) are full of selection bias, leading to a distribution shift from the collected graph training data to testing environments. We propose a novel stable prediction framework that permits \textit{locally} stable learning by capturing properties that are stable across environments in node representation learning, and \textit{globally} stable learning to balance the training of GNNs on different environments. We conduct extensive experiments on newly proposed and traditional graph benchmarks as well as datasets collected on real-world recommender systems concerning shift related to both node labels and attributes. Results demonstrate the capability of the proposed framework on stable prediction on graphs.

We believe that the insights of the proposed framework are inspirational to future developments of stable learning techniques on graphs. Mitigating selection biases can be essential for deploying GNNs on real-world application systems (\eg, confronting the inconsistencies between the training and testing distributions, and achieving fairness for different groups of users), while few works in the literature are proposed for learning stable GNNs. We plan to further investigate such a problem from the perspective of causal discovery/inference, \eg, disentangling the stable/unstable properties for a given neighborhood structure or a hidden variable. Another future direction is to design techniques for application-specific selection biases, \eg, exposure bias in recommender systems. We plan to make further investigations.

\bibliographystyle{ACM-Reference-Format}
\bibliography{sections/9.citations}

\balance

\end{document}